\documentclass[preprint,a4]{elsarticle}

\usepackage{amsmath,amssymb,amsfonts}
\usepackage{algorithmic}
\usepackage{graphicx}
\usepackage{textcomp}
\usepackage{xcolor}
\usepackage{lineno}
\usepackage[colorlinks=false,allbordercolors = {white}]{hyperref}	
\modulolinenumbers[5]

\biboptions{numbers,sort&compress} 
\usepackage{algorithm}
\usepackage{algorithmic}

\usepackage{multirow}
\usepackage{tikz}
\usepackage{pgfplots}
\usetikzlibrary{patterns}
\usetikzlibrary{calc}
\usepackage{arydshln}

\usepackage{adjustbox}

\journal{Pattern Recognition}

\newcommand{\LABTAB}[1]{\label{tab:#1}}
\newcommand{\LABFIG}[1]{\label{fig:#1}}
\newcommand{\LABSEC}[1]{\label{section:#1}}
\newcommand{\LABSSEC}[1]{\label{subsection:#1}}

\newcommand{\TAB}[1]{{Table~\ref{tab:#1}}}

\newcommand{\FIG}[1]{Fig.~\ref{fig:#1}} 

\newcommand{\SEC}[1]{Section~\ref{section:#1}}
\newcommand{\SSEC}[1]{Subsection~\ref{subsection:#1}}

\definecolor{darkred}{rgb}{.8,0,0}
\definecolor{brown}{rgb}{.5,.2,.4}

\newcommand{\notajj}[2]{{\color{blue}#2}} 

\newcommand{\notajc}[2]{{\color{blue}#2}} 
\newcommand{\x}{\mathbf{x}}
\newcommand{\dx}{\mathcal{X}}
\newcommand{\lnsk}{\vspace{.2cm}}
\newcommand{\lnup}{\vspace{-0.05cm}}

\begin{document}

\begin{frontmatter}

\title{Boosting offline handwritten text recognition in historical documents with few labeled lines}

\author[us]{José Carlos Aradillas}
\ead{jaradillas@us.es}
\author[us]{Juan José Murillo-Fuentes\corref{cor1}} 
\ead{murillo@us.es}
\author[uc3m,gmhri]{Pablo M. Olmos}
\ead{olmos@tsc.uc3m.es}

\cortext[cor1]{Corresponding author}

\address[us]{Dep. de Teoría de la Señal y Comunicaciones. Escuela Técnica Superior de Ingeniería.
Universidad de Sevilla. Camino de los Descubrimientos sn. 41092 Sevilla, Spain}
\address[uc3m]{Dep. de Teoría de la Señal y Comunicaciones,
 Universidad Carlos III de Madrid. Avda. de la Universidad 30. 28911 Leganes, Madrid,
Spain}
\address[gmhri]{Health Research Institute Gregorio Marañón, 28007 Madrid, Spain}


%
%

\begin{abstract}
In this paper, we face the problem of offline handwritten text recognition (HTR) in historical documents when few labeled samples are available and some of them contain errors in the train set. Three main contributions are developed. First we analyze how to perform transfer learning (TL) from a massive database to a smaller historical database, analyzing which layers of the model need a fine-tuning process. Second, we analyze methods to efficiently combine TL and data augmentation (DA). Finally, an algorithm to mitigate the effects of incorrect labelings in the training set is proposed. The methods are analyzed over the ICFHR 2018 competition database, Washington and Parzival. Combining all these techniques, we demonstrate a remarkable reduction of CER (up to 6\% in some cases) in the test set with little complexity overhead. 
\end{abstract}

\begin{keyword}
offline handwriting text recognition (HTR); connectionist temporal classification (CTC); historical documents; deep neural networks (DNN); convolutional neural networks (CNN);  long-short-term-memory (LSTM); outlier detection; transfer learning; data augmentation (DA).
\end{keyword}

\end{frontmatter}


\section{Introduction}\LABSEC{Intro}

The transcription of historical manuscripts is paramount for a better understanding of our history, as it allows for direct access to the contents, quite facilitating searches and studies. Also, classification and indexing of transcript text can be easily automated. 
Handwritten text recognition (HTR) tasks in historical datasets have been faced by many authors in the last few years \cite{Serrano10,Saabni2014,Yousefi15,Chammas2018,Strau18,Aradillas18,Soullard2019,Sanchez2019, Yousef2020, Aradillas20}. In HTR, transcribing each author can be considered a different task, since the distribution of both model input and output varies from writer to writer. 
At the input, we have variations not only in the calligraphy but also, depending on the digitization process, in the image resolution, contrast, color or background.
On the other hand, at the output the labels usually correspond to different languages and historical periods, with differences in the character set, the semantics and the lexicon. 

Usually, the process for automatic transcription of a document comprises 4 phases: 1) digitization of the document to obtain an image of every page in the document in electronic format; 2) segmentation of each page into corresponding regions with lines of text; 3) transcription of each line of text and finally 4) application of a dictionary or language model to correct errors in the transcription of texts as well as in the composition of the complete texts from the lines obtained in step 3).

While segmentation is an important issue in HTR \cite{BinMak2019, He2019, Saabni2014, Ares_Oliveira_2018}, this paper focuses on the transcription phase. In recent years, there has been a trend towards models based on deep neural networks (DNNs) \cite{Sanchez2019}. In particular, state of the art networks combine a convolutional neural network (CNN) \cite{Wu2017} with a recurrent neural network (RNN) with long-short-term-memory (LSTM) cells \cite{Hoch96}. This type of network models the conditioned probability, $p(l | \x)$, of a character sequence of arbitrary length, $l$, given an image, $ \x$, of fixed height and arbitrary width. Note that varying the number of characters yields a new design of the last layer of the DNN model.
These models are configured to minimize the connectionist temporal classification (CTC) cost function proposed by Graves in \cite{Graves06}.  
In some works 2D-LSTM \cite{Hoch96} networks are used \cite{Pham14, Voigtlaender16, Castro18, Moysset2018}. This RNN has two main drawbacks. On the one hand it has an extremely large number of parameters that makes learning difficult. On the other, it can not be parallelized \cite{Puigcerver17}. For these reasons it has been discarded here. 
Once the DNN model to be used has been designed, an enormous number of training samples are required to minimize the number of transcription errors, measured in character error rate (CER) or word error rate (WER), given by the Levenshtein distance \cite{Oncina2006} between the ground truth (GT) and the output of the model. However, in real problems we might only have a limited number of lines for a given author and document.  Besides, transcription of part of the documents to get labeled samples is expensive either in time or money. 
Take \cite{Serrano10} as an example, where  the manual transcription process of a document by a palaeography expert took an average of 35 minutes per page. In this scenario approaches allowing for a reduction in the transcript needed would very much improve the viability and cost of the process.
The goal of the methods proposed in this paper is to significantly reduce the number of annotated lines, thus reducing the monetary cost of the process. 
Contributions are threefold. First, we propose to use transfer learning (TL) by studying in detail which layers should be retained and which ones should be retrained. Then, we show that for a reduced number of lines, using data augmentation (DA) can be counterproductive. Some modifications are proposed to achieve improvements if both TL and DA are combined. Finally, we develop an approach to detect errors in the training database, i.e., characters that are wrongly transcripted and very much penalize the training of any DNN HTR system, particularly when a small database is available for training. The presence of unreliable labels is an issue we have found in some of the HTR databases. 
The remaining of the paper is organized as follows: in \SEC{RelatedWork} previous works reporting solutions to the problem of HTR over small datasets are summarized; in \SEC{Architecture}  the architecture and models used in this paper are presented;  in \SEC{TLandDA} the application of TL and DA for HTR is analyzed; in \SEC{CLP} an algorithm is proposed to detect and prone mislabeled lines in the training set; the paper ends with \SEC{Conclusions}, drawing main conclusions.


\section{Related Work}\LABSEC{RelatedWork}

In the HTR problem with a reduced training set, TL was applied by Soullard et al. in \cite{Soullard2019}. The main idea behind TL is initializing the parameters of a model by those learned from a huge dataset, denoted by \textit{source}. Then, the available labeled set of samples of the dataset of interest, the \textit{target}, is used to refine the parameters of the model, usually just a subset of them.  Hence, with TL we start learning a different task to avoid learning the whole set of parameters from scratch, preventing overfitting and favouring convergence. In \cite{Soullard2019}, they proposed a method which applies TL in both the optical and the language model. In this and other similar previous proposals on TL, the authors applied DA in both training and test steps. 


DA consists in augmenting the training set with synthetic generated samples. As TL, it reduces the tendency to overfitting when training models with a large number of parameters and limited labeled data. In DA for image classification problems, the training set is increased by modifying the original images through transformations such as scaling, rotation or flipping images, among others \cite{chatfield14}. There are several authors that have proposed specific DA techniques for HTR: in \cite{Wigington17} the authors apply methods for augmentation and normalization to improve HTR by allowing the network to be more tolerant of variations in handwriting by profile normalization. In \cite{Poznanski16} they show some affine transformation methods for data augmentation in HTR. In \cite{Krishnan16} and \cite{Shen16} they create new lines images by concatenating characters from different datasets. \cite{Krishnan16} does it from cursive characters while in \cite{Shen16} they do it from a database of handwritten Chinese characters. Similar to \cite{Wigington17}, in \cite{Simard2003} they also apply some elastic distortions to the original images. In \cite{Chammas2018} the authors improve the performance by augmenting the training set with specially crafted multi scale data. They also propose a model-based normalization scheme which considers the variability in the writing scale at the recognition phase. In these works they apply DA in relatively large well-known datasets, but here we show that the regularization effect of DA  technique has no impact when doing the fine-tuning adaptation to a singular writer in small databases. Accordingly, we conclude that the combination of TL and DA applied to small datasets has to be done carefully, to reduce the final error. 

Mislabeled detection in HTR has been seldom faced. In \cite{Nisa19} \notajc{where}{} they face a specific problem in the IAM database: crossed out words that are labeled with the symbol ``\textbf{\#}''. The authors propose a method in order to avoid how this specific label affects the performance. That method is focused in the specific problem of crossed out text and how it is annotated in the GT. The algorithm we propose in \SEC{CLP} is more general, addressing this and other found problems. In another related work in \cite{salah15}, the authors apply a method to align the output of a segmentation process with the available GT.

\section{Architecture and Databases}
\LABSEC{Architecture}

In the HTR pipeline, there are several aspects to improve the performance of a DNN model: preprocessing steps, architecture used, databases, regularization techniques, optimization, language model and dictionary, among others. The methods proposed in this paper are developed for an exemplary state-of-the-art DNN architecture, but can be easily included in the pipeline of any HTR system, reducing transcript errors. For a fair comparison, in this paper we use the same DNN model for all the experiments. Extra correction steps such as language model (LM) are not included but could be applied to further improve the performance. In this section we focus on the selection of the model of the DNN and the databases used.

\subsection{Architecture}
In this work, we implement a network architecture based on the convolutional recurrent neural network (CRNN) presented in \cite{Shi15}.
This proposal avoids the use of two dimensional LSTM (2D-LSTM) layers, applying convolutional layers as feature extractors and a stack of 1D bidirectional LSTM (BLSTM) layers to perform classification. Previous DNN architectures for HTR consisted of a combination of 2D-LSTM layers and convolutional layers, with a collapse stage before the output layer in order to reshape the features tensors from 2D to 1D \cite{Voigtlaender16, Pham14}. The use of 2D-LSTM layers at the first stages has several drawbacks such as the need for more memory in the allocation of activations and buffers during back-propagation and a larger runtime required to train the networks since parallel computation cannot be implemented as in CNN \cite{Puigcerver17}. Recently, it has been proven that CNN in the lower layers of an HTR system obtains similar features than a RNN containing 2D-LSTM units \cite{Puigcerver17}. 

\begin{figure*}[!t]
\centering
\includegraphics[width=4.9in]{{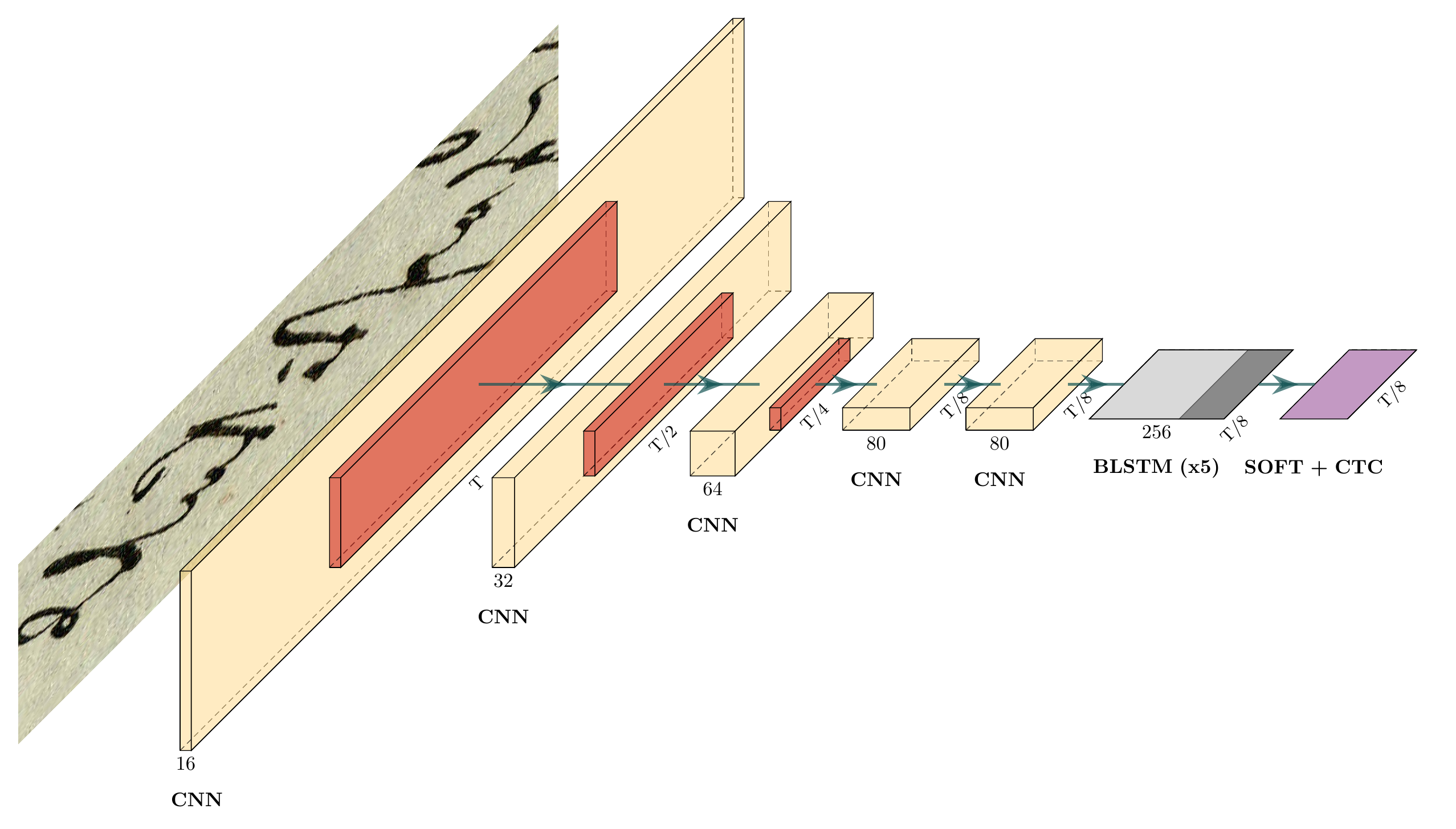}}%
\caption{The adapted CRNN architecture from \cite{Shi15} used as baseline.}
\LABFIG{structure}
\end{figure*}

The CRNN architecture proposed in \cite{Shi15} is comprised of seven CNN with max-pooling step at the output of four of them, followed by a stack of two BLSTM layers at the top of the network. In \cite{Aradillas18} we proved that the CRNN in \FIG{structure}, the one used in this work\footnote{Implementation is publicy available in https://github.com/josarajar/HTRTF}, achieves better performance than the original one proposed in \cite{Shi15}. 
It has a CNN with $5$ layers at the bottom, with a $3\times3$ and $1\times1$ stride kernel, the number of filters are $16$, $32$, $48$, $64$ and $80$, respectively. We use LeakyReLU as activation function. A $2\times2$ max-pooling is also applied at the output of the first $3$ layers, with the aim of reducing the size of the input sequence. At the output of the CNN, a column-wise concatenation is carried out with the purpose of transforming the 3D tensors of size $w \times h \times d$ (width $\times$ height $\times$ depth) into 2D tensors of size $w \times (h \times d$) where $w$ and $h$ are the width and height of the input image divided by 8, i.e., after 3 stages of $2\times2$ max-pooling. The depth, $d=80$, is the number of features of the last CNN layer. Therefore, at the output of the CNN we have sequences of length $w$ and depth $h\times80$ features.

After the CNN stage, $5$ 1D BLSTM recurrent layers of $256$ units without peepholes connections and hyperbolic tangent activation functions are applied. Since at the output of each BLSTM layer we have $256$ features in each direction, we perform a depth-wise concatenation in order to adapt the input of the next layer, in overall size of $512$. Dropout regularization \cite{Pham14, Srivastava14} is applied at the output of every layer, except for the first convolutional one, with rates $0.2$ for the CNN layers and $0.5$ for the BLSTM layers.

Finally, each column of features after the 5th BLSTM layer, with depth $512$, is mapped into the $L+1$ output labels with a fully connected network, where $L$ is the number of characters in the GT of each database, e.g., $79$, $83$, $96$ or $102$ in the IAM, Washington, Parzival or International Conference on Frontiers in Handwriting Recognition (ICFHR) 2018 Competition databases, respectively. 
 The additional dimension is needed for the blank symbol of the CTC \cite{Graves06}, that ends this architecture. Overall, this CNN-BLSTM-CTC architecture has a roughly number of to-be-learned parameters of $9.58\times10^6$, depending on the number of characters in each database.

\subsection{Databases}\LABSSEC{databases}
In this paper we focus on HTR over eight databases: IAM \cite{Marti02}, Washington \cite{Fisher12}, Parzival \cite{Fisher12}, and the five ones provided at the ICFHR 2018 Competition \cite{Strau18}.

\subsubsection{The IAM database}
The IAM database \cite{Marti02} contains $13353$ labeled text lines of modern English handwritten by $657$ different writers. The images were scanned at a resolution of $300$ dpi and saved as PNG images with $256$ gray levels. An image of this database is included in \FIG{IAMsample} alongside the GT transcript. The database is partitioned into training, validation and test sets of $6161$, $900$ and $2801$ lines, respectively\footnote{ The names of of the images of each set are provided in the \emph{Large Writer Independent Text Line Recognition Task}.}. Here, the validation and test sets provided are merged in a unique test set.
There are $79$ different characters in this database, including capital and small letters, numbers, some punctuation symbols and the white-space. 

\begin{figure}[htb]
\centering
\includegraphics[width=4.5in]{{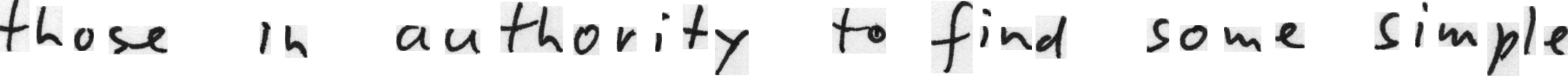}}%
\\
{\footnotesize GT: \textit{those in authority to find some simple} }
\caption{IAM handwritten text sample: image of a line and its transcript.}

\LABFIG{IAMsample}
\end{figure}

\subsubsection{The RIMES database}
The RIMES database is a collection of french letters handwritten by 1,300 volunteers who have participated to the RIMES database creation by writing up to 5 mails. The RIMES database thus obtained contains 12,723 pages corresponding to 5605 mails of two to three pages.
In our experiments, we take a set of 12111 lines extracted from the International Conference on Document Analysis and Recognition (ICDAR) 2011 line level competition. There are $100$ different characters in this database.

\subsubsection{The Washington database}
The Washington database contains $565$ text lines from the George Washington letters, handwritten by two writers in the 18th century. Although the language is also English, the text is written in longhand script and the images are binarized as illustrated in \FIG{Washingtonsample}, see \cite{Yousefi15} for a description of the differences between binarized and binarization-free images when applying HTR tasks. In this database four possible partitions \notajj{of the whole}{}are provided \notajj{in order}{}to train and validate. In this work we have randomly chosen one of them. The train, validation and test set contain $325$, $168$ and $163$ handwritten lines, respectively. There are $83$ different characters in the database.

\begin{figure}[htb]
\centering
\includegraphics[width=4.5in]{{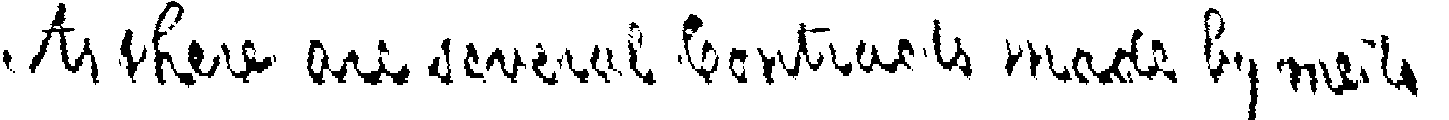}}%
\\
{\footnotesize GT: \textit{As there are several Contracts made by me to} }
\caption{Washington handwritten text sample: image of a line and its transcript.}

\LABFIG{Washingtonsample}
\end{figure}

\subsubsection{The Parzival database}
The Parzival database contains $4477$ text lines handwritten by three writers in the 13th century. In this case, the lines are binarized like in the Washington database, but the text is written in gothic script. A sample is included in \FIG{Parzivalsample}. There are $96$ different characters in this database. Note that the Parzival database has a large number of text lines in comparison to the Washington one.  We have randomly chosen a training set of the approximately same size as in the Washington training to emulate learning with a small dataset, the main goal of this work. 

\begin{figure}[bth]
\centering
\includegraphics[width=2in]{{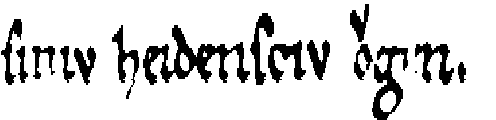}}%
\\
{\footnotesize GT: \textit{finiv heidenfciv ógen.}}
\caption{Parzival handwritten text sample: image of a line and its transcrit.}
\LABFIG{Parzivalsample}
\end{figure}

\subsubsection{The ICFHR 2018 Competition over READ dataset}
The set of documents of the ICFHR2018 Competition on Automated Text Recognition on a READ Dataset (https://readcoop.eu/) was proposed to compare the performance of approaches learning with few labeled pages. 
The dataset provided for the competition consists of 22 documents segmented at line level \cite{Strau18}. They are written in Italian and modern and medieval German. Each of them was written by only one writer but in different time periods and various languages. The training data is divided in a general set (of 17 documents) and a document-specific set (of 5 documents) called Konzilsprotokolle\_C, Schiller, Ricordi, Patzig and Schwerin of equal script as in the test set. 
Hereafter, \textit{general} is used to denote available {source} labeled databases different from the one of interest, while \textit{document-specific} denotes particular {target} documents. Also, the Konzilsprotokolle\_C dataset, of the University of Greifswald, will be abbreviated as Konzil.
The general database comprises roughly 25 pages per document (the precise number of pages varies such that the number of contained characters is almost equal per document). It will be denoted hereafter by ICFHR18-G.
For the 5 document-specific databases the authors provide 16 labeled pages plus 15 unlabeled pages. One can check for the error in the transcription of these databases by sending  the authors the transcription of these 15 pages. The results of the transcription are then published in the web of the contest.
 In \FIG{ICFHR2018sample}, samples from five specific target documents are displayed.
The standard Unicode Normalization Form Compatibility Decomposition - NFKD is applied to the GT to provide a common character set over such different documents, with 102 characters.
The goal of the competition is to fit a model to transcript each of the 5 specific target documents with the lowest CER possible, using the 17 source documents available for training. 
For each document-specific target dataset, four experiments are conducted, simulating that you have 0, 1, 4 or 16 annotated pages available for training.

\begin{figure}[tb!]
\centering
\includegraphics[width=4.5in]{{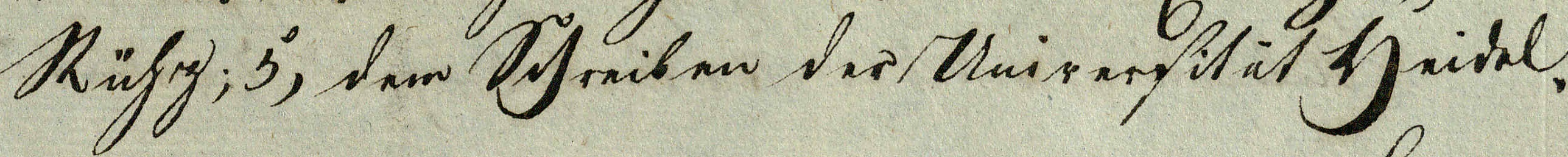}}\\
\lnup
{\footnotesize Konzil GT: \textit{Ruhz; 5, dem Schreiben der Universitat Heidel¬} }\\
\lnsk
\includegraphics[width=4.5in]
{{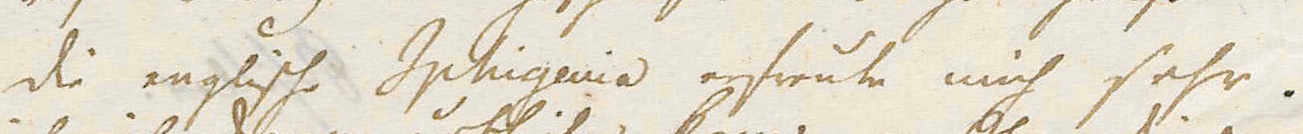}}\\
\lnup
{\footnotesize Schiller GT: \textit{Die englische Iphigenia erfreute mich sehr.} }\\
\lnsk
\includegraphics[width=4.5in]{{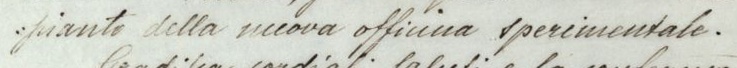}}\\
\lnup
{\footnotesize Ricordi GT: \textit{pianto della nuova officina sperimentale.} }\\
\lnsk
\includegraphics[width=4.5in]{{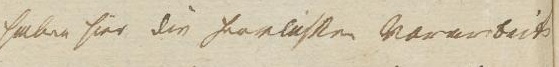}}\\
\lnup
{\footnotesize Patzig GT: \textit{haben hier die herrlichsten Vorarbeiten} }\\
\lnsk
\includegraphics[width=4.5in]{{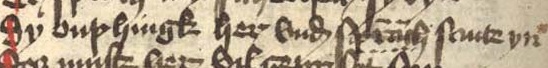}}\\
\lnup
{\footnotesize Schwerin GT: \textit{Dy onphingk her vnd  sante yn} }
\caption{From top to bottom: Konzil, Schiller, Ricordi, Patzig and Schwerin handwritten text samples with their transcripts.}
\LABFIG{ICFHR2018sample}
\end{figure}


\section{On the Data Augmentation and Transfer Learning Tradeoff}
\LABSEC{TLandDA}

As our first contribution, in this section we analyze the joint performance of TL and DA methods when applied to HTR\notajj{ tasks}{}. 

\subsection{Transfer learning}
\LABSEC{TransferLearn}

To cope with a reduced set of labeled inputs, we could first train the DNN model using as source available labeled large datasets. Then, we could apply TL or domain adaptation strategies \cite{Goodfellow2016} to tune the learnt model to transcript a target document.
As discussed in \SEC{Intro}, we usually deal with different tasks, where TL has proved useful to share the results of the learning between tasks. 

Formally, in HTR, deep learning algorithms have been usually focused on solving a problem over a domain  
$\mathcal{D} = \{\dx, P(\x)\}$, where $P(\x)$ is the marginal probability. Typically  $\x$ is the image for a segmented line in the text. The task consists of two components: a label space $\mathcal{Y}$  and an objective predictive function $f(\cdotp)$ (denoted by $\mathcal{T} = \{\mathcal{Y},f(\cdotp)\}$), which can be learned from the training data. The data consists of pairs $\{\x_i, y_i\}$, where $\x_i \in \dx$ and $y_i \in \mathcal{Y}$ \cite{Pan10} and $f(\x)=Q(y|\x)$ can be interpreted as the conditional probability distribution.

Given a source domain $\mathcal{D}_S$ and a learning task $\mathcal{T}_S$, transfer learning aims to help improve the learning of another target predictive function $f_T(\cdotp)$ in $\mathcal{D}_T$ using the knowledge in $\mathcal{D}_S$ and  $\mathcal{T}_S$. In this work we are interested in \emph{inductive transfer learning} in which the target task is different from the source task, as the domains are different ($\mathcal{D}_S\neq \mathcal{D}_T$). Here we perform TL by retraining a DNN model where 1) all weights are initialized to the ones of the DNN learned for $\mathcal{D}_S$ and  $\mathcal{T}_S$ and 2) the parameters of lowers layers can be fixed to the values of the ones obtained after training with other available source datasets, used as off-the-shelf feature extractors  \cite{Razavian14}.

In \cite{Aradillas18} we analyzed preliminary TL results over Washington and Parzival databases, by using the IAM database as source, and we investigated which layers should be kept fixed to then apply a fine tuning process to the others. 
We concluded that the best choice is to free all the layers, where the first one can be eventually fixed. 
In most cases, fixing only the first CNN layer leads to the best performance.

In \TAB{TLresults} we extend the analysis in \cite{Aradillas18} to the five specific documents in the ICFHR 2018 Competition dataset, where the 17 documents of the general set of the database, in the ICFHR18-G, are used as source. Results are included when fixing the layers 1 to 3 of the CNN, as fixing other layers provided larger errors in all cases. Lowest achieved errors are highlighted in boldface. Training set size is given in number of lines. It can be observed that, among all databases, the best performance is achieved when freeing all layers or, at most, only the first layer is kept fixed. Accordingly, hereafter the TL is applied by fixing just the first layer of the DNN model.
The results shown in all tables hereafter indicate mean values of CER or WER. To get the statistics, the model in \FIG{structure} is trained 10 times, where the parameters to initialize are independently and randomly set. In \TAB{TLresults}, a non-parametric bootstrapped confidence interval at 95\% \cite{Efron1987} is also included. For the remaining tables, the confidence intervals can be found in the supplementary material.

\begin{table}[htb]
\caption{Mean CER (\%) and bootstrapped confidence interval at 95\%, in brackets,
of the model in \FIG{structure} using TL for the Washington and Parzival target datasets  with the IAM database
as source and for Konzil, Schiller, Ricordi, Patzig and Schwerin datasets (see \SEC{Architecture}) as target domains with ICFHR18-G as source. The number of annotated lines used in training is included as `Train size'.}
\lnsk
\LABTAB{TLresults}
\begin{adjustbox}{width=\columnwidth,center}
\begin{tabular}{rc|rrrr}
\multicolumn{1}{l}{} & \multicolumn{1}{l|}{}                                        & \multicolumn{4}{c}{Fixed layers}                                                                                       \\ \hline
\multicolumn{1}{l}{} & \begin{tabular}[c]{@{}c@{}}Train \\size\end{tabular} & \multicolumn{1}{c}{All free} & \multicolumn{1}{c}{CNN 1} & \multicolumn{1}{c}{CNN 1,2} & \multicolumn{1}{c}{CNN 1,2,3} \\[2pt] \hline\hline
Washing.           & 325        & \textbf{5.3} [5.22-5.41]         & 5.4 [5.24-5.43]      & 5.5 [5.41-5.64]      & 6.3 [5.73-7.1]   \\[2pt] \hline
Parzival             & 350          & 3.3 [3.24-3.36]           & \textbf{3.3} [3.21-3.34]     & 3.5 [3.36-3.62]        & 3.6 [3.47-3.69]   \\[2pt] \hline\hline
Konzil               & 351             & 4.5 [4.33-4.61]      & \textbf{4.37} [4.24-4.54]      & 4.42 [4.36-5.49]          & 4.53 [4.43-4.61]      \\[2pt] \hline 
Schiller             & 238              & \textbf{9.4} [9.31-9.46]    & 9.42 [9.34-9.46]      & 9.48 [9.40-9.54]    & 10.1 [9.21-10.32]      \\[2pt] \hline
Ricordi              & 273             & 11.21 [11.14-11.25] & \textbf{11.2} [10.11-11.23]    & 11.28 [11.19-11.34] & 11.6 [11.41-11.72]   \\[2pt] \hline
Patzig               & 473                & 10.63 [10.51-10.70] & \textbf{10.6} [10.52-10.65]   & 10.68 [10.57-10.74]   & 12.4 [12.33-12.46]    \\[2pt] \hline
Schwerin             & 782            & \textbf{3.5} [3.46-3.53]        & \textbf{3.5} [3.47 - 3.51]  & 3.9 [3.81 - 3.94]   & 4.2 [4.15-4.26]      \\[2pt]        \hline
\end{tabular}
\end{adjustbox}
\end{table}

\subsection{Data augmentation}
\LABSEC{DataAug}

In \cite{Wigington17} the authors compare various DA approaches using both RIMES \cite{Fisher12} and IAM \cite{Marti02} databases as benchmarks, where transcription is made at word level.
Note that these databases have a considerably large number of labeled lines.
When not applying any augmentation technique, they get a CER of 5.35 \% (IAM) and 3.69 \% (RIMES).  The best CER values reported in \cite{Wigington17} by using various DA techniques are 3.93 \% and 1.36 \%, respectively. Which is equivalent to an improvement of approximately 2 \% in both databases.

Let us now extend the same analysis to scenarios with small training datasets: Washington, Parzival, Konzil, Schiller, Ricordi, Patzig and Schwerin databases.  
As throughout the paper, the transcriptions are made at line level. Results for the IAM, RIMES and the ICFGH18-G, i.e., the 17 documents of the general dataset in the ICFHR 2018 database, are also analyzed as references. In \TAB{AugmenatationComparison} we include the CER of our DNN model with no DA and two different DA techniques, affine transformation \cite{Poznanski16} and random warp grid distortion (RWGD) \cite{Wigington17}, for all databases in \SSEC{databases}.

In \TAB{AugmenatationComparison}, for the largest databases, the DA improvement is around 2\% (2\% in IAM, 1.9\% in RIMES and 2.5\% in ICFHR18-G). However, in the small databases, the CER reduction is remarkable, in the range 5 \% to 23.6 \%, see CERs highlighted in boldface.
Note that the results in \cite{Wigington17} are different to the ones in \TAB{AugmenatationComparison} because while in  \cite{Wigington17} transcription is done at the word level here whole lines are processed. This explains that
in IAM without DA we get CER 7.2\% while in \cite{Wigington17} a 5.35\% is reported. In any case,
it can be concluded that, since the DA acts as a regularization technique to avoid overfiting, the CER reduction is greater as the size of the training set is reduced.
At this point it is most interesting to compare the results of TL and DA, where it can be observed that TL exhibits, by far, the best CER reduction. Next, we face the design and analysis of both techniques combined. 

\begin{table}[htb]
\caption{Mean CER and WER (\%) with affine transformations \cite{Poznanski16} and RWGD\cite{Wigington17} DA approaches evaluated for all datasets in \SSEC{databases}.
The DNN is trained from scratch using the number of lines indicated by `Train size'. Largest DA CER reductions are highlighted in boldface.}
\lnsk
\LABTAB{AugmenatationComparison}
\begin{adjustbox}{width=.85\columnwidth,center}
\begin{tabular}{l|r|r|r|r|r|r|r}
\cline{2-8}
                                             & \multicolumn{1}{c|}{\multirow{2}{*}{\begin{tabular}[c]{@{}c@{}}Train size \end{tabular}}} & \multicolumn{2}{c|}{None}                           & \multicolumn{2}{c|}{\begin{tabular}[c]{@{}c@{}}Affine  Transf.\end{tabular}} & \multicolumn{2}{c}{\begin{tabular}[c]{@{}c@{}} RWGD\cite{Wigington17} \end{tabular}} \\
                                             & \multicolumn{1}{c|}{}                                                                                             & \multicolumn{1}{c|}{CER} & \multicolumn{1}{c|}{WER} & \multicolumn{1}{c|}{CER}                   & \multicolumn{1}{c|}{WER}                  & \multicolumn{1}{c|}{CER}                 & \multicolumn{1}{c}{WER}                 \\ \hline
\hline
\multicolumn{1}{l|}{RIMES}                  & 10163                                                                                                             & 4.4                      & 10.8                     & 2.7                                        & 10.7                                      & 2.5                                      & 10.4                                     \\ \hline  
\multicolumn{1}{l|}{IAM}                    & 6152                                                                                                              & 7.2                      & 22.2                     & 5.9                                        & 20.3                                      & 5.3                                      & 19.7                                     \\ \hline
\multicolumn{1}{l|}{Washington}             & 325                                                                                                               &  41.1                     & 85.3                     & 18.7                                       & 69.2                                      &  \textbf{17.5}                                     & 65.2                                     \\ \hline
\multicolumn{1}{l|}{Parzival}               & 350                                                                                                               & 18.2                     & 63.0                     & 14.1                                       & 56.4                                      & 12.9                                     & 53,6                                     \\ \hline\hline
\multicolumn{1}{l|}{ICFHR18-G} & 11424                                                                                                             & 12.2                     & 43.7                     & 10.6                                       & 40.1                                      & 9.7                                      & 38.6                                     \\ \hline
\multicolumn{1}{l|}{Konzil}     & 351                                                                                                               & 37.1                     & 95.4                     & 26.2                                       & 93.4                                      & 21.5                                     & 90.1                                     \\ \hline
\multicolumn{1}{l|}{Schiller}               & 238                                                                                                               & 45.4                     & 88.2                     & 32.4                                       & 87.6                                      & 30.1                                     & 85.5                                     \\ \hline
\multicolumn{1}{l|}{Ricordi}                & 273                                                                                                               & 57.2                     & 93.1                     & 36.2                                       & 91.1                                      & 35.2                                     & 90.3                                     \\ \hline
\multicolumn{1}{l|}{Patzig}                 & 473                                                                                                               & 24.5                     & 86.3                     & 18.5                                       & 81.3                                      & 17.1                                     & 80.5                                     \\ \hline
\multicolumn{1}{l|}{Schwerin}               & 782                                                                                                               &  21.1                    & 76.6                     & 17.4                                       & 72.9                                      & \textbf{16.5}                                     & 71.2                                     \\ \hline
\end{tabular}
\end{adjustbox}
\end{table}

\subsection{Combining data augmentation and transfer learning}
When comparing DA with TL, the large databases are excluded from the comparison. They play the role of source databases in the TL approach, specifically, the IAM is the source dataset when Washington and Parzival are targets and ICFHR18-G in the Konzil, Schiller, Ricordi, Patzig and Schwerin case. The RIMES database is only used in order to enhance the comparisons in this section.

In the combination of TL and DA techniques there are several possible designs. Here we propose the following two schemes.
In a first approach we perform DA at both the learning from the source dataset and the retraining of the model with the target one: 
\begin{enumerate}
\item{Train the model from scratch with a source dataset, applying DA.}
\item{Retrain the model with the target dataset, applying DA.}
\end{enumerate}
We name this proposal as DA-TL-DA. In a second proposal, denoted by DA-TL, no DA is applied to the target:
\begin{enumerate}
\item{Train the model from scratch with a source dataset, applying DA.}
\item{Retrain the model with the target dataset, \textbf{without} applying DA.}
\end{enumerate}

We perform the same experiments as in \SEC{TransferLearn}, obtaining the results included in \TAB{TL-DA-comparison-1}. In the first step of the DA-TL and DA-TL-DA methods, the model has been trained from scratch with the IAM database. After that, a fine tuning process is done over Parzival and Washington databases. In \TAB{TL-DA-comparison-2} the results are shown when training the model in \FIG{structure} from scratch with the ICFHR18-G, and being {fine-tuned} on the 5 specific target data sets provided. For the sake of completeness we include in \TAB{TL-DA-comparison-2}  the results for 0 pages in the target dataset, i.e., when no labeled sampled from the target is used. Note that in this case DA-TL-DA cannot be applied. 

\begin{table}[htb]
\centering
\caption{
Mean CER (\%) evaluated for Washington and Parzival datasets using TL and DA with IAM database as source. The number of annotated lines used in training is included as `Train size'. 
 }
\lnsk
\LABTAB{TL-DA-comparison-1}
\begin{adjustbox}{width=.8\columnwidth,center}
\begin{tabular}{lc|ccccc}
\hline
                                                & \multirow{2}{*}{\begin{tabular}[c]{@{}c@{}}Train size\\ \#lines \end{tabular}} & \multirow{2}{*}{None} & \multirow{2}{*}{TL} & \multicolumn{1}{l}{\multirow{2}{*}{DA}} & \multicolumn{1}{l}{\multirow{2}{*}{DA-TL-DA}} & \multicolumn{1}{l}{\multirow{2}{*}{DA-TL}} \\
    &                       &                       &                     & \multicolumn{1}{l}{}                    & \multicolumn{1}{l}{}                & \multicolumn{1}{l}{}            \\ \hline
\multicolumn{1}{c}{\multirow{3}{*}{Washington}} & 150        & 51.6         & 9.4        & 22.8          & 10.0           &   \textbf{9.3}     \\ \cdashline{2-7}   
\multicolumn{1}{c}{}                            		& 250       & 46.4         & 7.1        & 20.4           & 7.4             &   \textbf{7.0}   \\ \cdashline{2-7} 
\multicolumn{1}{c}{}                            		& 325       & 41.1         & 5.4        & 17.5          & 5.4             &  \textbf{5.4}      \\ \hline
\multirow{3}{*}{Parzival}                      		& 150        & 21.9        & 5.8      & 15.7           & 6.0               &   \textbf{5.6}   \\ \cdashline{2-7} 
                                                				& 250        & 20.7         & 4.0        & 14.2           & 4.2            &   \textbf{3.8}   \\ \cdashline{2-7} 
                                                				& 350       & 18.2         & 3.3        & 12.9           & 3.4            &   \textbf{3.3}   \\ \hline
\end{tabular}
\end{adjustbox}
\end{table}

In the light of \TAB{TL-DA-comparison-1} and \TAB{TL-DA-comparison-2}, it can be concluded that applying DA over the target training set once TL is applied, i.e., DA-TL-DA, either does not reduce the CER or even it slightly increases it, compared to the result of the TL approach alone or the DA-TL method. With the exception of Schwerin, in which DA-TL-DA slightly improves DA-TL.  Put in other words, in general, it is harmful to apply DA to the target dataset if TL has been applied, when just a reduced number of labeled lines are available in the target. On the other hand, DA+TL achieves improvements up to 5 \% in the ICFHR 2018 target documents, usually increasing with the reduction of the training set.

\begin{table}[htb]
\centering
\caption{Mean CER (\%) evaluated in ICFHR 2018 Competition Specific datasets as targets using TL and DA with ICFHR18-G as source. The number of annotated pages used in the training is included as `Train size'.}
\lnsk
\LABTAB{TL-DA-comparison-2}
\begin{adjustbox}{width=.85\columnwidth,center}
\begin{tabular}{lcccccc}
\hline
\multirow{2}{*}{}                           & \multicolumn{1}{c|}{\multirow{2}{*}{\begin{tabular}[c]{@{}c@{}}Training set \\ size. \# pages\end{tabular}}} & \multirow{2}{*}{None} & \multirow{2}{*}{TL} & \multicolumn{1}{l}{\multirow{2}{*}{DA}} & \multicolumn{1}{l}{\multirow{2}{*}{DA-TL-DA}} & \multicolumn{1}{l}{\multirow{2}{*}{DA-TL}} \\
                                            & \multicolumn{1}{c|}{}                                                                                           &                       &                     & \multicolumn{1}{l}{}                    & \multicolumn{1}{l}{}                & \multicolumn{1}{l}{}          \\ \hline
\multicolumn{1}{c}{\multirow{4}{*}{Konzil}} & \multicolumn{1}{c|}{0}         & --                    & 15.5                & --      & --\textbf{ }     &  \textbf{14.5}  \\ \cdashline{2-7} 
\multicolumn{1}{c}{}                        & \multicolumn{1}{c|}{1 {(29 lines)}}             & 48.1                  & 10.85      & 37.3         & 14.2      &    \textbf{10.8} \\ \cdashline{2-7} 
\multicolumn{1}{c}{}                        & \multicolumn{1}{c|}{4 {(116 lines)}}              & 45.3                  & 6.54       & 28.7        & 8.0         &  \textbf{6.51}  \\ \cdashline{2-7} 
\multicolumn{1}{c}{}                        & \multicolumn{1}{c|}{12 {(351 lines)}}      & 37.1                  & 4.37       & 21.5          & 5.0             & \textbf{4.32} \\ \hline
\multirow{4}{*}{Schiller}                   & \multicolumn{1}{c|}{0}     & --                    & \textbf{ 24.6}       & --              & --\textbf{}     &  \textbf{24.6}   \\ \cdashline{2-7} 
                                            & \multicolumn{1}{c|}{1 {(21 lines)}}            & 53.7                  & 17.36     & 39.5        & 21.4                   & \textbf{17.31} \\ \cdashline{2-7} 
                                            & \multicolumn{1}{c|}{4 {(84 lines)}}           & 48.4                    & 12.25     & 33.2           & 14.0            & \textbf{12.22} \\ \cdashline{2-7} 
                                            & \multicolumn{1}{c|}{12 {(238 lines)}}           & 45.4                  & 9.42      & 30.1           & 10.0                  &  \textbf{9.38}  \\ \hline
\multirow{4}{*}{Ricordi}                    & \multicolumn{1}{c|}{0}       & --                 & 39.19               & --              & --\textbf{ }        &     \textbf{34.2}   \\ \cdashline{2-7} 
                                            & \multicolumn{1}{c|}{1 {(19 lines)}}                & 56.2                  & 23.66     & 51.0      & 24.1             &   \textbf{22.71}\\ \cdashline{2-7} 
                                            & \multicolumn{1}{c|}{4 {(88 lines)}}                  & 43.5                  & 21.17               & 40.8        & 21.1    &  \textbf{21.02}   \\ \cdashline{2-7} 
                                            & \multicolumn{1}{c|}{12 {(273 lines)}}                & 37.2                  & 11.2                & 35.2         & \textbf{10.9}      &  11.1  \\ \hline
\multirow{4}{*}{Patzig}                     & \multicolumn{1}{c|}{0}         & --                    & 41.5                & --            & --\textbf{}      &   \textbf{38.2} \\ \cdashline{2-7} 
                                            & \multicolumn{1}{c|}{1 {(38 lines)}}                     & 42.5                  & 27.91     & 35.3        & 31.4           &   \textbf{26.7}     \\ \cdashline{2-7} 
                                            & \multicolumn{1}{c|}{4 {(156 lines)}}                   & 37.6                  & 16.4      & 30.5           & 18.3            & \textbf{16.1}     \\ \cdashline{2-7} 
                                            & \multicolumn{1}{c|}{12 {(473 lines)}}                & 24.5                  & 10.6       & 17.1       & 11.2                 & \textbf{10.0}  \\ \hline
\multirow{4}{*}{Schwerin}                   & \multicolumn{1}{c|}{0}      & --                    & 34.5                & --         &--\textbf{ }          &  \textbf{31.3}  \\ \cdashline{2-7} 
                                            & \multicolumn{1}{c|}{1 {(68 lines)}}              & 38.4                  & 12.15               & 30.2         & \textbf{10.6}         &  10.8  \\ \cdashline{2-7} 
                                            & \multicolumn{1}{c|}{4 {(264 lines)}}                & 29.3                  & 5.73                & 24.3       & \textbf{5.3}            & 5.5 \\ \cdashline{2-7} 
                                            & \multicolumn{1}{c|}{12 {(782 lines)}}             & 21.1                  & 3.5                 & 16.5          & \textbf{3.3}           &  3.4\\ \hline
\end{tabular}
\end{adjustbox}
\end{table}

From the discussion above, and bearing \TAB{TL-DA-comparison-1} and \TAB{TL-DA-comparison-2} in mind, it can be concluded that DA-TL is a robust approach. When fine-tuning a DNN that have been previously trained with a similar task (a huge database of HTR samples), the starting point is reasonably good as we can observe in \TAB{TL-DA-comparison-2} when the training set size is 0 pages. Afterwards, the DNN model is trained with the target database. 
Only a few samples are available in the target set, that
represent just a limited part of the support of the its marginal distribution, $P_T(\x)$. 
After TL, the parameters of the DNN encode information from both the source and the target training sets. At this point, we conjecture that by using DA in the target dataset and further re-fining the parameters, the DNN model overfits to the augmented versions of the target samples, forgetting the knowledge learnt from the source one, that very much helps to transcript inputs out of the support generated by augmenting the target set.   
This leads to an increase in the final CER. 

\section{The Corrupted Label Purging (CLP) Algorithm}
\LABSEC{CLP}
In this section, we focus on the impact in the learning of the DNN model of the number and quality of the lines in the target datase.
We first analyze the impact in the performance of the number of \emph{healthy} lines, i.e., lines with no transcription errors in the training dataset. Then we study how this performance degrades with label errors. Finally, we propose an algorithm to detect and remove potential label errors in the dataset.

\subsection{Sensitivity to the number of transcript lines} 

When a small number of lines is available in the target training set, deep learning models are quite sensitive to a small variation in the number of annotated lines. In this subsection this sensitivity is evaluated in a specific dataset from the ICFHR 2018 Competition \cite{Strau18}. The chosen training dataset consists of 16 pages from the Konzil, segmented at line level.

The ICFHR 2018 target datasets have 16 labeled pages each. Unless otherwise indicated, hereafter 4 of them will be used for testing purposes while up to 12 pages will be used for training.
Usually a 10 \% out of the used training set is devoted to validation. The ICFHR18-G dataset is used as source database in the TL-DA approach.

\begin{figure}[tb]
\centering

\begin{tikzpicture}
\hspace{.18cm}
\begin{axis}[  
    grid = both,
    width=3.5in,
    height=2in,
    ylabel near ticks,
    xlabel near ticks,
    ymin=4, ymax=13, xmin=28, xmax=350,
    xlabel={{\#} of lines ($l$) in the training set },
    ylabel={{CER \% }},
   name=ax1,
   legend style={at={(0.98,0.85)},nodes={scale=0.8, transform shape}}]
\addplot[mark=x,color=blue] table [x=Lines, y=CER, col sep=comma] {img/KonzilSABS.csv};
\addplot table [x=Lines, y=CER, col sep=comma] {img/KonzilSAedited.csv};
\addlegendentry{Original training set}
\addlegendentry{Training set with errors}

  \coordinate (c1) at (axis cs:28,8);
  \coordinate (c2) at (axis cs:28,12.8);
  \draw (c1) rectangle (axis cs:60,12.8);

\end{axis}

\begin{axis}[
   name=ax2,
   scaled ticks=false,
   xmin=28,xmax=60,
   ymin=8,ymax=12.8,
  at={($(ax1.south east)+(0.3in,5)$)},
   grid=both,
       width=1.6in,
    height=1.8in
 ]
\addplot[mark=x,color=blue] table [x=Lines, y=CER, col sep=comma] {img/KonzilSABS.csv};
\addplot table [x=Lines, y=CER, col sep=comma] {img/KonzilSAedited.csv};

\end{axis}

\draw [dashed] (c1) -- (ax2.south west);
\draw [dashed] (c2) -- (ax2.north west);

\end{tikzpicture}
\\
(a)
\vspace{.5cm}

\begin{tikzpicture}
\begin{semilogyaxis}[
    grid = both,
    width=4in,
    height=2in,
    ylabel near ticks,
    xlabel near ticks,
    ymin=0.006, ymax=0.3, xmin=28, xmax=350,
    xlabel={{\#} of lines ($l$) in the training set },
    ylabel={$ \frac{\bigtriangleup CER \%}{\bigtriangleup l} $},
    legend style={nodes={scale=0.8, transform shape}}]
\addplot[mark=x,color=blue] table [x=Lines, y=CER, col sep=comma] {img/KonzilSABS_d.csv};
\addplot table [x=Lines, y=CER, col sep=comma] {img/KonzilSAedited_d.csv};
\addlegendentry{Original training set}
\addlegendentry{Training set with errors}
\end{semilogyaxis}
\end{tikzpicture}
\\
{(b)}
\\
\caption{ 
 (a) CER  (\%) divided by the number of annotated lines, $l$, used and (b) decrement of CER  (\%) divided by the number of new labeled lines added to obtain it, $\Delta l$, 
in the training of the DNN model with DA-TL approach using the ICFHR18-G dataset as source and the Konzil dataset as target with no artificial errors ({\color{blue}$\times$}) corrupted with artificial errors ({\color{darkred}$\blacksquare$}).}

\LABFIG{SensitivityCurve}
\end{figure}

In \FIG{SensitivityCurve}.(a) the blue curve in {\color{blue}$\times$} represents the TL-DA CER versus the available number of lines, $l$, of the target training set in the range 29-350 lines, corresponding to 1 and 12 pages, respectively. In the left part of the figure, the CER decreases at a rate of 1\% every 4 new lines added to the training set. After approximately 50 lines, the decreasing rate of the CER changes to approximately 1\% every 100 lines. This is evidenced in \FIG{SensitivityCurve}.(b) where it is depicted the absolute value of the variation of the CER (\%), $\Delta CER$, with the increment of the number of annotated lines used in the target to achieve it, $\Delta l$.
 It can be concluded that the sensitivity to the number of samples in the training set is significantly larger for small training sets.

In  \FIG{SensitivityCurve} we also include the ``Training set with errors'' curve ({\color{darkred}$\blacksquare$}) which corresponds to the analysis above but where labeling errors have been artificially introduced, as follows. The annotation of a line is modified with probability $L$. Then, within a modified labeling, a character is changed with probability $R$. In both cases following a Bernoulli disrtibution. Every changed character is replaced by any independently and randomly selected character, following a discrete uniform probability.
In \FIG{SensitivityCurve}, where $L=0.2$ and $R=0.3$, it is interesting to note that the impact of labeling errors in the CER value is more dramatic for small training sets while the rate at which the CER decreases with the number of lines added remains roughly unaltered.

\subsection{Types of transcription errors}
\LABSEC{TypesTE}

Prior to propose approaches to detect mislabels in the training set, we discuss on three typical types and causes of errors in the datasets, as follows.

\begin{enumerate}
\item\textit{{Mislabeled characters.}}
When labeling a training set, the most common mistake is to confuse a character with another, usually a similar one. This can be seen in the well-known IAM database \cite{Marti02}, where in the labels it is indicated that some lines could have some annotation errors. This type of error is the one simulated in \FIG{SensitivityCurve}.

\item\textit{{Label Misalignment.}}
The second kind of detected errors is due to a misalignment in the labels. This could be caused by, e.g., a mistake in the name given to some images in the database. This error is encountered several times in the Ricordi dataset from the ICFHR 2018 Competition \cite{Strau18} as illustrated in \FIG{Misalingment}. It can be observed in this example that the transcript does not correspond to the handwritten text in the image above. On the contrary, it is quite close to the model output, after being trained with several lines of the dataset.


\begin{figure}[h]
\centering
\includegraphics[width=4.5in]{{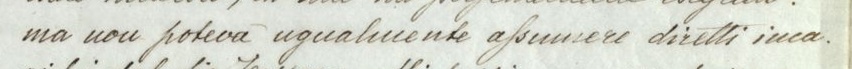}}%
\\
{\footnotesize GT: \textit{meno d’osservarle che cio non e corretto: in ogni}. \\Model output: \textit{ma non poteva nqualuiente assunere direlti inca¬}\vspace{-.1cm}}
\caption{Sample of a completely mislabeled text at Ricordi dataset.}
\LABFIG{Misalingment}
\end{figure}

\begin{figure}[!t]
\centering
\includegraphics[width=4.5in]{{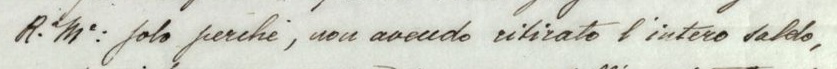}}%
\\
{\footnotesize GT: \textit{R\textbf{[icchezz]}a M\textbf{[obil]}e solo perche, non avendo ritirato l’intero saldo}. \\Model output: \textit{N.° Mi: solo perche, non avendo ritirato l’intero saldo}\lnup}
\caption{Sample of special annotations in the GT at the Ricordi dataset.}
\LABFIG{Brackets}
\end{figure}

\item\textit{{Special annotations in the ground truth.}}
Perhaps, the most common source of error is due to special annotations that some transcribers or database managers introduce in some datasets to include some notes inline. In \cite{Nisa19} they found this problem in the IAM database: crossed out words that are labeled with the symbol ``\textbf{\#}" followed by the word behind the blot. Training the model with this labeling might lead to an unpredictable behaviour, since the model could replace the text using ``\textbf{\#}" at different parts of the text. The model will either be able to recognize the text behind the blot or replace the word by the symbol ``\textbf{\#}", or both. Another special annotation is included in \FIG{Brackets}, where they write in brackets extra characters that are not in the handwritten text. The output of a model trained with samples of the same dataset is showed below the GT. Despite in this line the CER is about 35\%, it can be observed that the model output is quite similar to the handwritten text. 
\end{enumerate}


Manually annotating historical documents remains a challenging task that is prone to errors, even for experts in the field. As discussed in the previous section, when a huge set of annotated samples is available, deep learning models do not suffer from some mislabeled samples, as they better generalize. However, when a limited set of annotated lines of a specific writer is available to train, mislabeled lines induce an overfitting to transcripts with errors, quite hard to tackle via regularization. In the example in \FIG{SensitivityCurve} we illustrate this problem, when just a few mislabeled lines are introduced.

\begin{algorithm}[htb]
\begin{algorithmic}
\STATE 
{\bf Given inputs}: source set $x_S\in\mathcal{X}_S$ and $y_S \in \mathcal{Y}_S$, target set $x_T \in \mathcal{X}_T$ and $y_T \in \mathcal{Y}_T$ and threshold, $\epsilon$.
\vspace{0.3cm}
\STATE 1) Fit the prediction function $f_S(y_S|x_S)$ with the {source} training set $\{x_{S}, y_{S}\}$.
\vspace{-0.4cm}
\STATE 2)  Split the {target} training set into $N$ subsets $\{x_{T_1}, x_{T_2},...,x_{T_N}\}$,
$\{y_{T_1}, y_{T_2},...,y_{T_N}\}$ .
\vspace{-0.4cm}
\STATE
\FOR {$n=1,...,N$}
\STATE
3) Initialize the prediction function $f_{T_n}(\cdot) = f_S(\cdot)$.
\STATE
4) Fine tune the prediction function with the whole target set except for the $n${th}, $\{x_{T_i \neq T_n}\}$, $\{y_{T_i\neq T_n}\}$.
\\
\STATE
5) Include in the new target set, $\{x_{T'}, y_{T'}\}$,  all pairs $\{x_{T_{n}}^{(i)}, y_{T_{n}}^{(i)}\}$ whose predictions $f_{T_n}(y_{T_n}^{(i)}|x_{T_n}^{(i)})$ have errors below a CER threshold, $\epsilon$.
\ENDFOR 
\STATE
6) Initialize the prediction function $f_{T}(\cdot) = f_S(\cdot)$.
\STATE
7) Fine tune the prediction function, $f_{T'}(y_{T'}|x_{T'})$, to the modified target set $\{x_{T'}, y_{T'}\}$.
\\
\vspace{0.3cm}
{\bf Output}:
\STATE
Function $f_{T}(y_{T}|x_{T})$ over the target domain $\mathcal{D}_T$.
\end{algorithmic}
\caption{Corrupted Labels Purging (CLP)}\label{alg:uno}
\end{algorithm}

\subsection{Mislabel detection algorithm} 
As one of our main contributions, we propose an algorithm to detect and remove mislabeled lines from the training set, detailed in Algorithm \ref{alg:uno}. A block diagram of the algorithm is also depicted in \FIG{Algorithm}. It divides the target training dataset into $N$ subsets. For every subset, $n$, the method performs DA-TL using the rest of subsets, $k=1,\hdots,N$, $k\neq n$, as training set and it evaluates the CER metric over the subset $n$. Lines with CER above a threshold, $\epsilon$, in the $n$th subset are detected as wrongly transcribed and discarded. Hence, we are implementing some sort of $k$-fold validation, in which the size of each validation partition is reduced after removing problematic lines. Finally, the DA-TL is applied to the resulting target database.   
In \FIG{Hist} we include the histogram of the CER per line for the 5 ICFHR 2018 document-specific datasets using the CLP algorithm with $N=2$. The ICFHR18-G was used as source. The histograms were estimated with the CER of the outputs of the $n=1,2$ stages computed with the lines not used in the training, see the output of ``Target subset $n$'' blocks in \FIG{Algorithm}.
In the left column, models have been trained with 4 pages while in the right column they have been trained with 12 pages. Lines are corrupted with artificial errors with probability $L=0.1$, while every character in the label of a line is changed with probability $R=0.3$ to a random value.
Conservatively, we believe that a 10\% average number of corrupted lines represent a label error rate similar to the one we encounter in real databases. It is interesting to observe that the results for the Schwerin dataset are remarkably better than for the others, because it has a significantly larger number of lines per page.  Besides, in the Ricordi dataset, the histogram for 12 pages exhibits a large values around 0.8. This dataset is known to have label misalignments.

\begin{figure*}[!t]
\centering
\includegraphics[width=5in]{{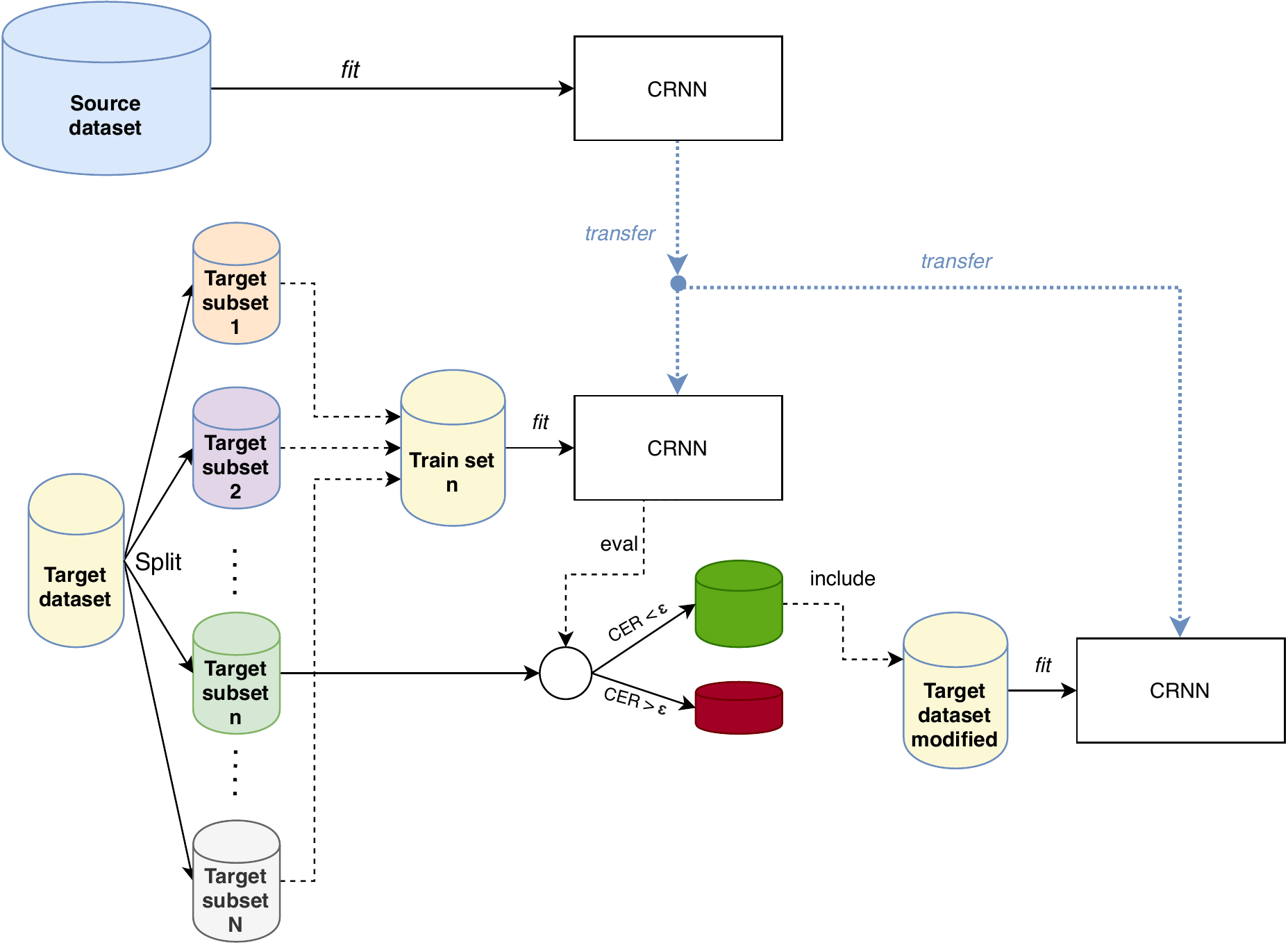}}%
\\
\caption{Corrupted labels purging algorithm. The algorithm applied over target subset $n$ is depicted. The same procedure should be applied to all the subsets to build the \textit{Target dataset modified}.}
\LABFIG{Algorithm}
\end{figure*}

\begin{figure}[htbp]
\centering
\begin{tikzpicture}
\begin{axis}[
    grid = both,
    width=2.4in,
    height=1.4in,
    axis lines=middle,
    axis line style={->},
    ylabel near ticks,
    xlabel near ticks,
    xmin=0, xmax=1,
    xlabel={CER },
    ylabel={{\#} lines}]
\addplot [ybar,draw = blue,
                           fill=blue!50, bar width=1]
                 table [x=CER, y=Nlines, col sep=comma] {img/Konzil4.csv};
\addplot +[mark=none, dashed, thick, red] coordinates {(0.5, 0) (0.5, 7)};
\node[below right, align=center, text=black] at (rel axis cs:0.28,0.6) {$76\%$};
\addplot +[mark=none, dashed, thick, red] coordinates {(0.7, 0) (0.7, 12)};
\node[below right, align=center, text=black] at (rel axis cs:0.48,0.9) {$84\%$};
\end{axis}
\end{tikzpicture}
\begin{tikzpicture}
\begin{axis}[
    grid = both,  
    width=2.4in,
    height=1.4in,
    axis lines=middle,
    axis line style={->},
    ylabel near ticks,
    xlabel near ticks,
    xmin=0, xmax=1,
    xlabel={CER },
    ]
\addplot [ybar,draw = blue,
                           fill=blue!50, bar width=1]
                 table [x=CER, y=Nlines, col sep=comma] {img/Konzil12.csv};
\addplot +[mark=none, dashed, thick, red] coordinates {(0.5, 0) (0.5, 30)};
\node[below right, align=center, text=black] at (rel axis cs:0.28,0.6) {$92\%$};
\addplot +[mark=none, dashed, thick, red] coordinates {(0.7, 0) (0.7, 50)};
\node[below right, align=center, text=black] at (rel axis cs:0.48,0.95) {$96\%$};                
\end{axis}
\end{tikzpicture}

\vspace{-.1cm}
(a) Konzil

\begin{tikzpicture}
\begin{axis}[
    grid = both,    
    width=2.4in,
    height=1.4in,
    axis lines=middle,
    axis line style={->},
    ylabel near ticks,
    xlabel near ticks,
    xmin=0, xmax=1,
    xlabel={CER },
    ylabel={{\#} lines}]
\addplot [ybar,draw = blue,
                           fill=blue!50, bar width=1]
                 table [x=CER, y=Nlines, col sep=comma] {img/Schiller4.csv};
\addplot +[mark=none, dashed, thick, red] coordinates {(0.5, 0) (0.5, 6)};
\node[below right, align=center, text=black] at (rel axis cs:0.28,0.7) {$79\%$};
\addplot +[mark=none, dashed, thick, red] coordinates {(0.7, 0) (0.7, 9)};
\node[below right, align=center, text=black] at (rel axis cs:0.48,0.95) {$89\%$};
\end{axis}
\end{tikzpicture}
\begin{tikzpicture}
\begin{axis}[  
    grid = both,  
    width=2.4in,
    height=1.4in,
    axis lines=middle,
    axis line style={->},
    xlabel near ticks,
    ymin=0, ymax=30, xmin=0, xmax=1,
    xlabel={CER },
    ]
\addplot [ybar,draw = blue,
                           fill=blue!50, bar width=1]
                 table [x=CER, y=Nlines, col sep=comma] {img/Schiller12.csv};
\addplot +[mark=none, dashed, thick, red] coordinates {(0.5, 0) (0.5, 18)};
\node[below right, align=center, text=black] at (rel axis cs:0.28,0.6) {$90\%$};
\addplot +[mark=none, dashed, thick, red] coordinates {(0.7, 0) (0.7, 25)};
\node[below right, align=center, text=black] at (rel axis cs:0.48,0.9) {$95\%$};  
\end{axis}
\end{tikzpicture}

\vspace{-.1cm}
(b) Schiller

\begin{tikzpicture}
\begin{axis}[  
    grid = both,  
    width=2.4in,
    height=1.4in,
    axis lines=middle,
    axis line style={->},
    ylabel near ticks,
    xlabel near ticks,
    xmin=0, xmax=1,
    xlabel={CER },
    ylabel={{\#} lines}]
\addplot [ybar,draw = blue,
                           fill=blue!50, bar width=1]
                 table [x=CER, y=Nlines, col sep=comma] {img/Ricordi4.csv};
\addplot +[mark=none, dashed, thick, red] coordinates {(0.5, 0) (0.5, 7)};
\node[below right, align=center, text=black] at (rel axis cs:0.28,0.7) {$$71\%};
\addplot +[mark=none, dashed, thick, red] coordinates {(0.7, 0) (0.7, 11)};
\node[below right, align=center, text=black] at (rel axis cs:0.48,1) {$87\%$};
\end{axis}
\end{tikzpicture}
\begin{tikzpicture}
\begin{axis}[  
    grid = both,  
    width=2.4in,
    height=1.4in,
    axis lines=middle,
    axis line style={->},
    xlabel near ticks,
    ymin=0, ymax=30, xmin=0, xmax=1,
    xlabel={CER },
    ]
\addplot [ybar,draw = blue,
                           fill=blue!50, bar width=1]
                 table [x=CER, y=Nlines, col sep=comma] {img/R12.csv};
\addplot +[mark=none, dashed, thick, red] coordinates {(0.5, 0) (0.5, 18)};
\node[below right, align=center, text=black] at (rel axis cs:0.28,0.7) {$83\%$};
\addplot +[mark=none, dashed, thick, red] coordinates {(0.7, 0) (0.7, 28)};
\node[below right, align=center, text=black] at (rel axis cs:0.48,1) {$88\%$};  
\end{axis}
\end{tikzpicture}

\vspace{-.1cm}
(c) Ricordi

\begin{tikzpicture}
\begin{axis}[  
    grid = both,  
    width=2.4in,
    height=1.4in,
    axis lines=middle,
    axis line style={->},
    ylabel near ticks,
    xlabel near ticks,
    xmin=0, xmax=1,
    xlabel={CER },
    ylabel={{\#} lines}]
\addplot [ybar,draw = blue,
                           fill=blue!50, bar width=1]
                 table [x=CER, y=Nlines, col sep=comma] {img/Patzig4.csv};
\addplot +[mark=none, dashed, thick, red] coordinates {(0.5, 0) (0.5, 9)};
\node[below right, align=center, text=black] at (rel axis cs:0.28,0.7) {$85\%$};
\addplot +[mark=none, dashed, thick, red] coordinates {(0.7, 0) (0.7, 13)};
\node[below right, align=center, text=black] at (rel axis cs:0.48,1) {$90\%$};
\end{axis}
\end{tikzpicture}
\begin{tikzpicture}
\begin{axis}[  
    grid = both,  
    width=2.4in,
    height=1.4in,
    axis lines=middle,
    axis line style={->},
    xlabel near ticks,
    xmin=0, xmax=1,
    xlabel={CER },
    ]
\addplot [ybar,draw = blue,
                           fill=blue!50, bar width=1]
                 table [x=CER, y=Nlines, col sep=comma] {img/Patzig12.csv};
\addplot +[mark=none, dashed, thick, red] coordinates {(0.5, 0) (0.5, 28)};
\node[below right, align=center, text=black] at (rel axis cs:0.28,0.6) {$87\%$};
\addplot +[mark=none, dashed, thick, red] coordinates {(0.7, 0) (0.7, 45)};
\node[below right, align=center, text=black] at (rel axis cs:0.48,1) {$94\%$};  
\end{axis}
\end{tikzpicture}

\vspace{-.1cm}
(d) Patzig
\\
\begin{tikzpicture}
\begin{axis}[  
    grid = both,  
    width=2.4in,
    height=1.4in,
    axis lines=middle,
    axis line style={->},
    ylabel near ticks,
    xlabel near ticks,
    xmin=0, xmax=1,
    xlabel={CER },
    ylabel={{\#} lines}]
\addplot [ybar,draw = blue,
                           fill=blue!50, bar width=1]
                 table [x=CER, y=Nlines, col sep=comma] {img/Schwerin4.csv};
\addplot +[mark=none, dashed, thick, red] coordinates {(0.5, 0) (0.5, 55)};
\node[below right, align=center, text=black] at (rel axis cs:0.28,0.6) {$96\%$};
\addplot +[mark=none, dashed, thick, red] coordinates {(0.7, 0) (0.7, 100)};
\node[below right, align=center, text=black] at (rel axis cs:0.42,1) {$100\%$};
\end{axis}
\end{tikzpicture}
\begin{tikzpicture}
\begin{axis}[  
    grid = both,  
    width=2.4in,
    height=1.4in,
    axis lines=middle,
    axis line style={->},
    xlabel near ticks,
    xmin=0, xmax=1,
    xlabel={CER },
    ]
\addplot [ybar,draw = blue,
                           fill=blue!50, bar width=1]
                 table [x=CER, y=Nlines, col sep=comma] {img/Schwerin12.csv};
\addplot +[mark=none, dashed, thick, red] coordinates {(0.5, 0) (0.5, 150)};
\node[below right, align=center, text=black] at (rel axis cs:0.28,0.6) {$98\%$};
\addplot +[mark=none, dashed, thick, red] coordinates {(0.7, 0) (0.7, 230)};
\node[below right, align=center, text=black] at (rel axis cs:0.42,1) {$100\%$};  
\end{axis}
\end{tikzpicture}

(e) Schwerin

\caption{Histogram of CER with DA-TL and ICFHR18-G as source dataset for the 5 document-specific datasets using 4 pages (left) and 12 pages (right) of the target dataset. Lines and characters were corrupted with probabilites $L=0.1$ and $R=0.3$ respectively. The histograms were evaluated with the outputs of the $N=2$ target subsets. Red dashed lines indicate the percentage of lines with CER$\leq \epsilon$ with $\epsilon =50\%$ and $\epsilon =70\%$, left and right lines, respectively. }
\LABFIG{Hist}
\end{figure}
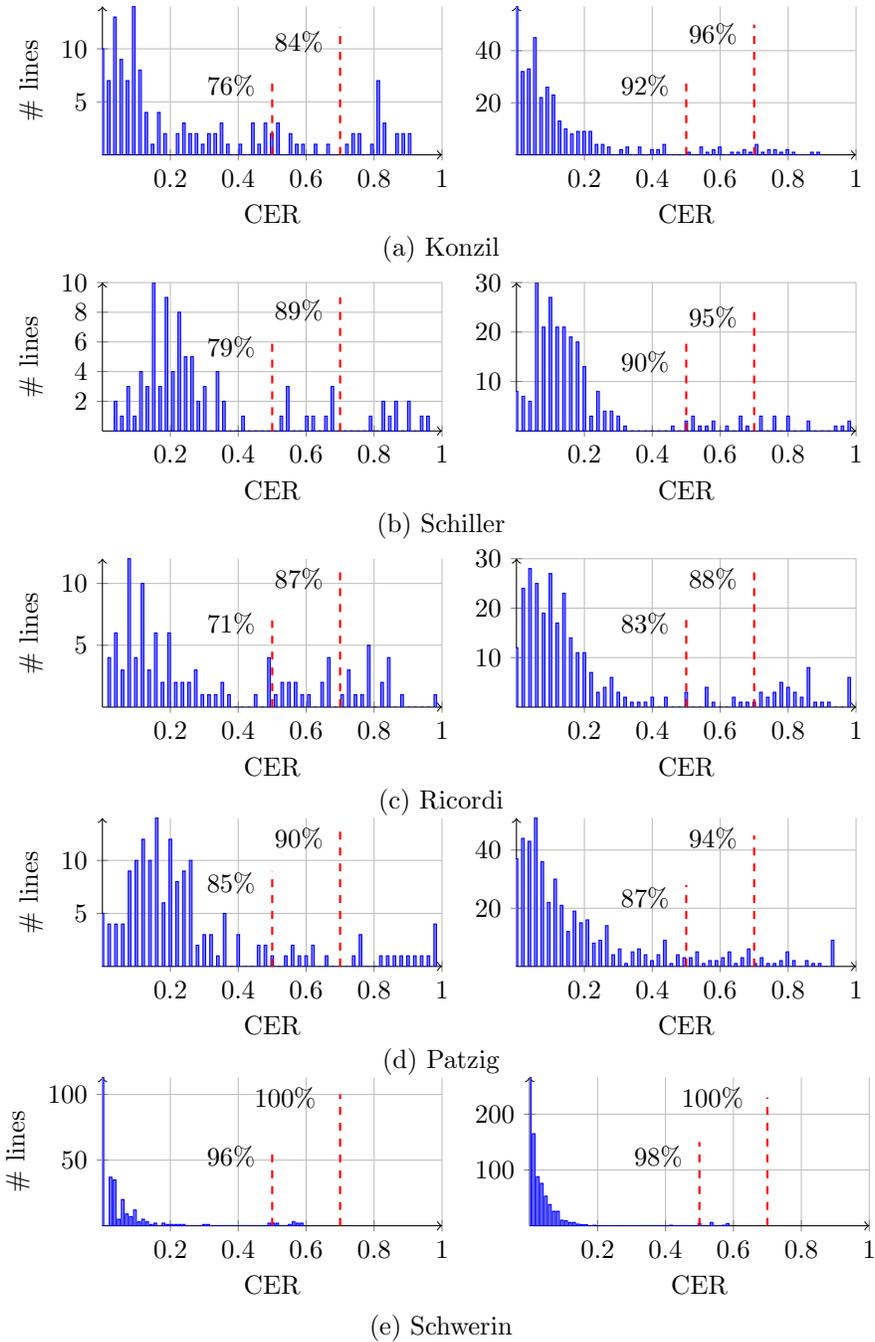

\subsection{CLP threshold analysis}
\LABSSEC{CLPasessment}

The selection of the threshold is central to the algorithm performance. In \FIG{Hist} the CER of the healthy lines is mainly distributed around a mode value, to the left of each histogram, while outliers exhibit larger values. As representative values to be studied, after extensive simulations, we restrict our analysis to the thresholds $\epsilon = 0.5$ and $\epsilon = 0.7$, for an average rate $L=0.1$ of artificially modified lines, and $R=0.3$.
In \FIG{Hist} we indicate the percentage of lines with CER equal or lower than $0.5$ and $0.7$, left and right red dashed lines in the subfigures, respectively. We conclude that almost 10\% of lines have a CER above $\epsilon = 0.7$ when 4 pages for training are available and the same occurs in the case of 12 pages when $\epsilon = 0.5$. 

The selection for $\epsilon$ should not lead to the deletion of healthy lines, otherwise the overall CER would raise. On the other hand, the threshold must ensure a  sensitivity when corrupted lines are encountered.  

In the following, we study the CLP algorithm in two different scenarios. The first experiment we perform consists in applying the CLP algorithm to the ICFHR 2018 target databasets, with 4 and 12 pages as target training set size. Then we evaluate the CLP for the Washington and Parzival databases, with 150 and 325 lines as target training set size. 
The same procedure is followed through all the scenarios:
\begin{enumerate}
\item Fit the model to the source set.
\item {Run} DA-TL plus CLP.
\end{enumerate}

\subsubsection{ICFHR 2018 Competition Results}\LABSSEC{dumb}
We test the CLP algorithm over real databases where we do not have any prior knowledge about the pattern of labeling errors. We do also include artificial errors to evaluate the CLP robustness.

The results of these analyses are reported in  \TAB{ICFHR2018t} and \TAB{WashingtonParzival}.
Their three last columns include the results for the DA-TL with no CLP as `Baseline', for the DA-TL+CLP with $\epsilon=50\%$ and then for the DA-TL+CLP with $\epsilon=70\%$. For every target dataset and training set size three rows are used to report the CER (\%) when no artificial errors are introduced, $R=0$, for $R=30$\% and $R=50$\%.

In this first case the ICFHR18-G dataset is used as source. The 17 documents of this corpus has a total number of 11424 lines. The DA-TL plus CLP was applied to the five target documents in the competition:  Konzil, Schiller, Ricordi, Patzig and Schwerin. The results are included in \TAB{ICFHR2018t}, where it is included the average value for the CER and the number of removed lines by the CLP algorithm. 

\begin{table}[tbh!]
\centering
\caption{Mean CER (\%) evaluated in Konzil, Schiller, Ricordi, Patzig and Schwerin target documents in the ICFHR2018 Competition datasets for DA-TL, DA-TL+CLP with $\epsilon=50\%$ and $\epsilon=70\%$. DA-TL was applied with both a training set of 4 pages and 12 pages. The annotation for a line is corrupted with probability $L=0.1$ and a character within it randomly replaced with probability $R$. $R=0$ indicates no error introduced in the labelings. The number of removed lines by the CLP algorithm are included in parentheses in the last two columns. Best achieved value in every row is in boldface.}
\lnsk
\begin{adjustbox}{width=.85\columnwidth,center}
\LABTAB{ICFHR2018t}
\begin{tabular}{c |c |c |l |l |l  }

\multicolumn{1}{l|}{ Dataset}  &\multicolumn{1}{|l|}{Train set size}     & $R$ & Baseline & $\epsilon=50$\% & $\epsilon=70$\% \\ \hline
\multirow{6}{*}{Konzil} &
\multirow{3}{*}{\begin{tabular}[c]{@{}c@{}}4 Pages \\ (116 lines)\end{tabular}}  & 0\%                   & \textbf{7.6}               & 8.5(-31)         & 7.9(-7)          \\  
  & & 30\%               & 8.7              & 8.3 (-41)        & \textbf{7.82} (-14)        \\ 
     &  & 50\%                  & 9.1                & 8.2 (-39)        & \textbf{7.9} (-16)        \\ \cdashline{2-6}
& \multirow{3}{*}{\begin{tabular}[c]{@{}c@{}}12 Pages \\ (351 lines)\end{tabular}} & 0\%                   & \textbf{4.6 }              & 5.3 (-1)         & 4.7 (-0)         \\   
& & 30\%               & 5.3               & \textbf{4.6} (-29)        & 5.0 (-25)        \\  
&     & 50\%                  & 5.5               & \textbf{4.8} (-35)        &  5.0(-28)        \\ \hline
\multirow{6}{*}{Schiller} & \multirow{3}{*}{\begin{tabular}[c]{@{}c@{}}4 Pages \\ (84 lines)\end{tabular}}  & 0 \%&  \textbf{13.27} & 14.72 (-12)  & 13.61 (-5)\\  
 &   & 30 \%& 15.19 & 14.81 (-17) & \textbf{14.43} (-10)\\  
&  & 50 \%& 15.64 & 14.96 (-22) & \textbf{13.87} (-12)\\ \cdashline{2-6} 
& \multirow{3}{*}{\begin{tabular}[c]{@{}c@{}}12 Pages \\ (244 lines)\end{tabular}} & 0 \%&  \textbf{9.42} & 9.76 (-2)  & 9.42 (-0)\\  
 & & 30 \%& 11.31 & \textbf{10.41} (-22) & 10.62 (-22)\\        
& & 50 \%& 12.75 & 10.61 (-24) & \textbf{10.51} (-25)\\       \hline
\multirow{6}{*}{Ricordi} & \multirow{3}{*}{\begin{tabular}[c]{@{}c@{}}4 Pages \\ (88 lines)\end{tabular}}  & 0 \%&  21.1 & 18.2 (-16) & \textbf{18.2} (-16) \\   
& & 30 \%& 23.2 & 20.8 (-32) & \textbf{20.5} (-27) \\  
& & 50 \%& 24.31 & 21.94 (-44) & \textbf{20.81} (-27) \\\cdashline{2-6} 
   & \multirow{3}{*}{\begin{tabular}[c]{@{}c@{}}12 Pages \\ (295 lines)\end{tabular}} &0 \%&  9.7 & \textbf{9.4} (-38)  & 9.4 (-38) \\  
& & 30 \%& 10.47 & \textbf{9.23} (-41) & 9.49 (-38) \\        
& & 50 \%& 10.8 & \textbf{9.53} (-52) & 9.75 (-44) \\       \hline
\multirow{6}{*}{Patzig} &\multirow{3}{*}{\begin{tabular}[c]{@{}c@{}}4 Pages \\ (156 lines)\end{tabular}}  & 0 \%&  \textbf{18.3} & 18.93 (-7)  & 18.32 (-0)\\  
& & 30 \%& 21.41 &  21.6 (-27) &  \textbf{21.1} (-18)\\  
& & 50 \%& 21.84 & 22.12 (-27) & \textbf{21.31} (-18)\\ \cdashline{2-6} 
& \multirow{3}{*}{\begin{tabular}[c]{@{}c@{}}12 Pages \\ (473 lines)\end{tabular}} & 0 \%&  \textbf{11.5} & 11.96 (-15)  & 11.54 (-4)\\   
&  & 30 \%& 12.28 & 12.23 (-61) & \textbf{11.98 (-52)}\\        
&  & 50 \%& 12.8 & 12.67 (-63) & \textbf{12.35 (-54)}\\       \hline
\multirow{6}{*}{Schwerin} &\multirow{3}{*}{\begin{tabular}[c]{@{}c@{}}4 Pages \\ (264 lines)\end{tabular}}  & 0 \%&  5.3 & 5.3 (-0)  & \textbf{5.3} (-0)\\  
& & 30 \%& 5.36 &  \textbf{5.31} (-14) &  5.36 (-0)\\ 
&  &50 \%& 5.39 & \textbf{5.32} (-26) & 5.33 (-12)\\  \cdashline{2-6}
& \multirow{3}{*}{\begin{tabular}[c]{@{}c@{}}12 Pages \\ (782 lines)\end{tabular}} & 0 \%&  3.3 & 3.3 (0)  & \textbf{3.3} (0)\\  
& & 30 \%& 3.36 &  \textbf{3.31} (-14) &  3.36 (-0)\\     
& & 50 \%& 3.53 & \textbf{3.34} (-75) & 3.39 (-22)\\      \hline
\end{tabular}
\end{adjustbox}
\end{table}

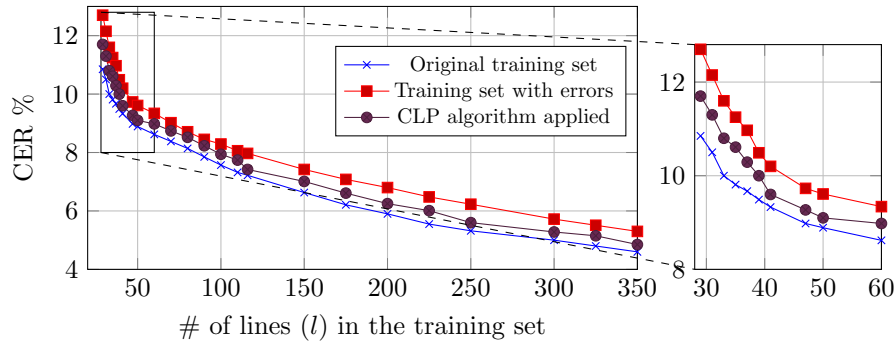
\begin{figure}[tb]
\centering

\begin{tikzpicture}
\hspace{.18cm}
\begin{axis}[  
    grid = both,
    width=3.5in,
    height=2in,
    ylabel near ticks,
    xlabel near ticks,
    ymin=4, ymax=13, xmin=20, xmax=350,
    xlabel={{\#} of lines ($l$) in the training set },
    ylabel={{CER \% }},
    legend style={at={(0.98,0.85)},nodes={scale=0.8, transform shape}}]
\addplot[mark=x,color=blue] table [x=Lines, y=CER, col sep=comma] {img/KonzilSABS.csv};
\addplot table [x=Lines, y=CER, col sep=comma] {img/KonzilSAedited.csv};
\addplot table [x=Lines, y=CER, col sep=comma] {img/KonzilSAcorrected.csv};
\addlegendentry{Original training set}
\addlegendentry{Training set with errors}
\addlegendentry{CLP algorithm applied}
  \coordinate (c1) at (axis cs:28,8);
  \coordinate (c2) at (axis cs:28,12.8);
  \draw (c1) rectangle (axis cs:60,12.8);
\end{axis}

\begin{axis}[
   name=ax2,
   scaled ticks=false,
   xmin=28,xmax=60,
   ymin=8,ymax=12.8,
  at={($(ax1.south east)+(0.3in,0)$)},
   grid=both,
       width=1.6in,
    height=1.8in,
 ]
\addplot[mark=x,color=blue] table [x=Lines, y=CER, col sep=comma] {img/KonzilSABS.csv};
\addplot table [x=Lines, y=CER, col sep=comma] {img/KonzilSAedited.csv};
\addplot table [x=Lines, y=CER, col sep=comma] {img/KonzilSAcorrected.csv};
\end{axis}

\draw [dashed] (c1) -- (ax2.south west);
\draw [dashed] (c2) -- (ax2.north west);

\end{tikzpicture}

\caption{ CER  (\%) divided by the number of annotated lines, $l$, with the DA-TL approach using the ICFHR18-G dataset as source and the Konzil dataset as target with no artificial errors ({\color{blue}$\times$}), with artificial errors ({\color{darkred}$\blacksquare$}) and with artificial errors and CLP used ({\color{brown}$\bullet$}).}

\LABFIG{SensitivityCurveWithCLP}
\end{figure}

In the view of the results we highlight the following aspects.
First note that, when errors are induced, the threshold $\epsilon=70\%$ performs better in most of the cases when the training set is of 4 pages while the threshold $\epsilon=50\%$ is the best choice for 12 pages. Exceptions can be observed in Patzig and Schwerin corpora. For the Patzig dataset we conclude that $\epsilon=70\%$ is the best choice in any case. This is due to the distribution of the errors in this dataset, that has a larger variance and therefore more lines are above the $\epsilon=50\%$ CER , it can be seen in \FIG{Hist}. In the Schwerin corpus, the threshold 50\% has the best CER in all cases, the opposite that in the Patzig dataset. This is due to the distribution of the errors in this dataset that, due to the larger number of lines used, has a lower variance and most lines are below the 10\% CER (see \FIG{Hist}). 

It is also interesting to remark that  in the Ricordi case, the algorithm improves the CER in the original dataset, i.e., without synthetic errors. This is explained by the fact that in this dataset, as already discussed, there are some mislabeled lines like in the case illustrated in \FIG{Misalingment}.\notajc{In \SEC{MisalingmentCorrection} we use this corpus as a certain target in which we test the algorithm without enforcing synthetic errors.}{}
Besides, note that for $R=0$ and $\epsilon = 70\%$  a large number of removed lines is quite an indicator of the dataset containing errors in the annotated lines.

For the sake of completeness we include in \FIG{SensitivityCurveWithCLP} the evolution of the CER versus the number of lines, $l$, in \FIG{SensitivityCurve} including the CER for the proposed algorithm (CLP) ($\circ$). The introduction of the CLP improves the TL-DA approach when the dataset has corrupted lines. In the range $l=[40,50]$ the TL-DA with CLP with $l=40$ achieves the same CER as the DA-TL with $l=50$ lines in the training set. 

\subsubsection{Washington and Parzival Results}
In this second analysis, the model is pre-trained with the IAM database as the source dataset to train the model with DA-TL for the Washington and Parzival targets. There are two main differences to the previous study of the ICFHR 2018 datasets: 1) the number and set of characters are different from the source and targets datasets and 2) we compare the CER of both targets in terms of the number of lines instead of the number of pages, where we consider two cases, 150 lines and 325 lines, similar to the number of lines used in the previous scenario.

First rows in \TAB{WashingtonParzival} include the results obtained after fine tuning the model to the Washington dataset. In this study the threshold $\epsilon=70\%$ is the best option when the number of lines is 150 while $\epsilon=50\%$ exhibits the lowest CER when the number of lines is 325. This is equivalent to the 4 and 12 pages in the Konzil, Schiller and Ricordi cases in which the number of lines is similar.  For these thresholds: we get an improvement of 0.8 and 0.63 in the case of 150 lines and no deterioration over the original dataset. In the case of 325 lines we get a boost of 0.4 and 0.5 and no deterioration over the original dataset.

Results obtained after fine tuning the model to the Parzival dataset are also included in \TAB{WashingtonParzival}, see the lower rows. 
Similar conclusions can be drawn except for $R=30\%$ and 150 lines, where the 50\% exhibits the best CER. If we choose the threshold as in the previous cases, 70\% and 50\%, we still get a slight improvement or at least, no deterioration.


\begin{table}[tb!]
\centering
\caption{Mean CER (\%) evaluated in Washington and Parzival documents for DA-TL, CLP with threshold $\epsilon=50\%$ and CLP with threshold $\epsilon=70\%$. DA-TL was applied with the IAM dataset as source and using 150 and 325 lines from the target. 
The annotation for a line is corrupted with probability $L=0.1$ and a character within it randomly replaced with probability $R$. $R=0$ indicates no error introduced in the labelings. The number of removed lines by the CLP algorithm are included in parentheses in the last two columns.}
\lnsk
\begin{adjustbox}{width=.85\columnwidth,center}
\LABTAB{WashingtonParzival}
\begin{tabular}{c |c |c |l |l |l }

\multicolumn{1}{l|}{Dataset}  &\multicolumn{1}{|l|}{Train set size}     & R & Baseline & \textbf{$\epsilon=50$\%} & \textbf{$\epsilon=70$\%} \\ \hline
\multirow{6}{*}{Washington} &
\multirow{3}{*}{\begin{tabular}[c]{@{}c@{}} 150 lines\end{tabular}}  & 0 \%& 9.4  & 9.5 (-6) & \textbf{9.4} (-2)\\   
& & 30 \%& 11.3& 10.6 (-20)&  \textbf{10.5}(-14)\\  
& &50 \%& 11.5& 11.1 (-31) &  \textbf{10.87}(-19)\\ \cdashline{2-6} 
& \multirow{3}{*}{\begin{tabular}[c]{@{}c@{}} 325 lines \end{tabular}} & 0 \%& 5.3  & \textbf{5.3} (-2) & 5.3 (-0)\\ 
&  & 30 \%& 6.1& \textbf{5.7} (-26)& 6.1 (-0)\\      
& & 50 \%& 6.3& \textbf{5.8} (-34)& 6.3 (-0)\\      \hline    

\multirow{6}{*}{Parzival} & \multirow{3}{*}{\begin{tabular}[c]{@{}c@{}} 150 lines\end{tabular}}  & 0 \%& 5.8  & 5.8 (-0) & \textbf{5.8} (-0)\\    
& & 30 \%& 6.4 & \textbf{6.0} (-15)& 6.2 (-2)\\    
& & 50 \%& 6.6 & 6.2 (-20) &  \textbf{6.1}(-14)\\  \cdashline{2-6}  
& \multirow{3}{*}{\begin{tabular}[c]{@{}c@{}} 325 lines\end{tabular}} & 0 \%& 3.3  & \textbf{3.3} (-0) & 3.3 (-0)\\     
& & 30 \%& 3.5  & \textbf{3.5} (-0)& 3.5 (-0)\\          
& & 50 \%& 3.5 & \textbf{3.4} (-35)& 3.5 (-0)\\      \hline                 
                                                                                        
\end{tabular}
\end{adjustbox}
\end{table}

\subsection{Correcting label misalignment}
\LABSEC{MisalingmentCorrection}

In \SEC{TypesTE} we summarized the different types of transcription errors. 
One of these errors is due to misalignment of the annotations with the images. When a high number of lines are classified as mislabels, this type of error can be addresed by searching within the outputs of the DNN model for the whole target dataset, the transcript best fitting every annotation in the GT, hence aligning annotations and images in the dataset. This approach is quite similar to the one proposed in \cite{salah15}.

In the case of the Ricordi dataset in the ICFHR 2018 competition we realized that the CLP detected a high number of mislabeled lines in the dataset. Note the large numbers of removed lines in \TAB{ICFHR2018t} for $\epsilon=70\%$ and this dataset with $R=0$. By simply visual inspection we confirmed that the error was of the type of misalignment of images and annotations. Here, we apply the CLP plus the simple automatic alignment approach described above.  

The comparison between simply removing the mislabeled lines and correcting the alignment of the database, is shown in \TAB{RicordiCorrected}. In this table one can observe a significant dropping in the CER when correcting these misalignments of the lines. In the training with 4 pages, the overall decrease is of 3.7\%. In the 12 pages analysis, the CER drops 0.3 \% when removing the lines while it further decreases 0.8 \% when correcting them. Note also that the gain is higher when a lower number of annotated lines are used.

\begin{table}[bt]
\caption{Comparison between the CLP algorithm with line removal and the CLP plus alignment of the GT after detection. The mean CER (\%) is evaluated for the Ricordi document with a training set of size 4 pages (88 lines) and 12 pages (295 lines).}
\lnsk
\LABTAB{RicordiCorrected}
\begin{adjustbox}{width=0.8\columnwidth,center}
\begin{tabular}{c |c |c c c}

\multicolumn{1}{l|}{\textbf{Train set size}}                                             & \textbf{Method} & \textbf{Baseline} & \textbf{$\epsilon=50$\%} & \textbf{$\epsilon=70$\%} \\ \hline
\multirow{2}{*}{\textbf{\begin{tabular}[c]{@{}c@{}}4 pages \\ (88 lines)\end{tabular}}}  & CLP &  21.1 & \textbf{18.2}  & \textbf{18.2} \\ \cdashline{2-5} 
                                                                                          & CLP + alignment& 21.1 & \textbf{17.4}  & \textbf{17.4} \\  \hline
\multirow{2}{*}{\textbf{\begin{tabular}[c]{@{}c@{}}12 pages \\ (295 lines)\end{tabular}}} &CLP &  9.7 & \textbf{9.4}   & 9.4 \\ \cdashline{2-5} 
                                                                                          & CLP + alignment & 9.7 & \textbf{8.9}   & 8.9  \\       \cline{2-5}  \hline
\end{tabular}
\end{adjustbox}
\end{table}


\section{Conclusions}
\LABSEC{Conclusions}

In this paper we analyze, for small training sets and in the framework of historical HTR, two well-known techniques in almost every deep learning application: TL and DA. We show that TL improves the CER between 10-40\% when applied to small training sets, of the order of 300 text lines. DA also drops the CER in a range of 2-20\% when a network is trained from scratch. In TL, DA in the source dataset does reduce the CER. However, applying DA to the target dataset jointly with TL exhibits worse results than using TL alone.  Hence, we propose the DA-TL approach where the DA is applied to the source dataset in the TL proccess. 

Besides, we highlight that the DNN models are very sensitive to the number of lines in the train set when this number is low. Therefore, errors in annotated lines of small target dataset have a greater impact than in large datasets, for the same proportions of mislabels. To avoid that, we propose a method which can detect the mislabeled lines and remove it from the training set. Furthermore, we propose to fix errors of the misalignment type, by searching for the true labels in the datasets.


By using the proposed 5+5 DNN model with CNN and BLSTM layers followed by a CTC, we conclude analyzing the results of the novel DA-TL approach over the ICFHR 2018 Competition\footnote{ The results are publicly available in the ICFHR competition website: https://scriptnet.iit.demokritos.gr/competitions/10/viewresults/}. The results included in \TAB{Results} were reported by the organisers of the competition. The contestants provide the transcript of the 15 test pages for every document in the target set: Konzil, Schiller, Ricordi, Patzig and Schwerin. Then, the organizers evaluate the CER, publicly publishing the results. In this table our results are compared against the 5 original contestants in the competition: OSU \cite{Wigington17}, ParisTech \cite{Chammas2018}, LITIS \cite{Swaileh2017} , PRHLT and RPPDI.  These approaches use DNN models based on CNN, LSTM and CTC, where some variant of the LSTM is used. Some of them use DA in the target, and LM. The recent work published by Yousef et al. \cite{Yousef2020} using a DNN model based on a fully gate convolutional network (GCN), outperformed the LSTM based approaches, with a mean value of $13.02$ \% providing a $23.35$ \% CER for a 0 page training size. 

The results of the proposal in this work are included in the lowest rows of \TAB{Results} where, following the conclusions in \SSEC{CLPasessment}, we used $\epsilon=70\%$ for the 1 and 4 pages training and $\epsilon=50\%$ for the 16 pages. Also, the CLP includes an alignment stage.
Results are presented in three groups of columns. First, the average CER (\%) for the 5 target dataset is included when 0, 1, 4 and 16 pages of the target datases are used. The second group of 5 columns report the average CER (\%) for the learning with 0, 1, 4, and 16 pages in the dataset for every document. The mean value per row is included in the last column.

\begin{table}[]
\caption{CER ICFHR 2018 Competition results for LSTM based models: upper part, other previous approaches and, in the lower part, the results for the approaches in this work. Lowest mean values in both parts are highlighted in boldface.}
\lnsk
\LABTAB{Results}
\begin{adjustbox}{width=\columnwidth,center}

\begin{tabular}{l|llll|lllll|l}
               & \multicolumn{4}{c|}{CER (\%) per  training size}
               & \multicolumn{5}{c|}{CER (\%) per document}        & Mean    \\ \hline
               & 0                         & 1                        & 4                        & 16                       & Konzil  & Schiller & Ricordi & Patzig  & Schwerin &         \\ \hline
OSU\cite{Wigington17}           & 31.40                   & 17.74                  & 13.27                  & 9.02                   & 9.39  & 21.10  & 23.27 & 23.17 & 12.98  & \textbf{17.86} \\
ParisTech\cite{Chammas2018}     & 32.25                   & 19.80                  & 16.98                  & 14.72                  & 10.49 & 19.05  & 35.60 & 23.83 & 17.02  & 20.94 \\
LITIS\cite{Swaileh2017}          & 35.30                   & 22.51                  & 16.89                  & 11.34                  & 9.14  & 25.69  & 30.50 & 25.18 & 18.04  & 21.51 \\
PRHLT          & 32.79                   & 22.15                  & 17.89                  & 13.33                  & 8.65  & 18.39  & 35.07 & 26.26 & 18.65  & 21.54 \\
RPPDI\cite{Moysset2018}         & 30.80                   & 28.40                  & 27.25                  & 22.85                  & 11.90 & 21.88  & 37.29 & 32.75 & 28.55  & 27.32 \\ \hline
TL     & 32.77       & 19.51                  & 15.12                  & 8.26                   & 9.16  & 21.00  & 29.39   & 23.25 & 13.54  & 18.93 \\
DA-TL & 31.55                   & 19.21                  & 14.91                  & 8.16                   & 8.58  & 21.68  & 27.84 & 22.35 & 12.50  & 17.83 \\
CLP     & 30.13                 & 19.10                  & 12.40                  & 7.93                   & 8.59  & 21.69  & 22.81 & 22.35 & 12.51  & \textbf{17.39}
\end{tabular}

\end{adjustbox}
\end{table}

It can be observed that the DA-TL and CLP outperforms all approaches within the CNN+LSTM+CTC class hence underlining the importance of the issues discussed: DA is important but in the source dataset, TL is to be considered and mislabeling detection and correction is important if the dataset exhibits errors. Besides, the CLP introduces a residual $0.01\%$ loss if the datasets have no errors in the labels while the reduction is important if they have, see the results for the Ricordi corpus where a reduction of $6.58$ \% is achieved. The presence of errors in this database was detected by checking the number of removed lines by the CLP.

At this point, it is interesting to mention that other variations of the algorithm have been tried to further improve the performance. In this sense, we tried to evaluate the CTC loss \cite{Graves06}  to select a threshold $\epsilon$. We found it complex to deal with because it depends on several factors as the number of epochs in the training or if batch normalization has been applied. In future work, we expect to improve the algorithm in this way. Another promising research line could be introducing TL-DA and CLP in other DNN models, such as the based on GCN \cite{Yousef2020}, that has a quite low value for 0 pages, to further improve the CER. Besides, introducing LM in the proposed DA-TL and CLP approaches could be also investigated.

\section*{Acknowledgment}
Funding: This work was partially supported by the Spanish government MEC [grant numbers FPU16/04190, project MINECO TEC2016-78434-C3-2-3-R]; the Comunidad de Madrid [grant numbers IND2017/TIC-7618, IND2018/TIC-9649, and Y2018/TCS-4705]; the BBVA Foundation [Deep-DARWiNproject]; and the European Union (FEDER and the European Research Council (ERC) through the European Union Horizon 2020 research and innovation program under [grant number 714161]).


\bibliography{mybibfile}

\end{document}